\documentclass{bmvc2k}

\usepackage{graphicx}
\usepackage{amsmath}
\usepackage{amssymb}
\usepackage{booktabs}
\usepackage{makecell}
\usepackage{subfigure}

\title{Stating Comparison Score Uncertainty and Verification Decision Confidence Towards Transparent Face Recognition}

\addauthor{Marco Huber}{marco.huber@igd.fraunhofer.de}{1,2}
\addauthor{Philipp Terh\"{o}rst}{philipp.terhoerst@uni-paderborn.de}{1,3,4}
\addauthor{Florian Kirchbuchner}{florian.kirchbuchner@igd.fraunhofer,de}{1}
\addauthor{Naser Damer}{naser.damer@igd.fraunhofer.de}{1,2}
\addauthor{Arjan Kuijper}{arjan.kuijper@igd.fraunhofer.de}{1,2}

\addinstitution{
 Fraunhofer Institute for Computer Graphics Research IGD \\ Darmstadt, Germany\\
}
\addinstitution{
Technical University of Darmstadt \\ Darmstadt, Germany\\
}

\addinstitution{
Norwegian University of Science and Technology \\ Gj{\o}vik, Norway
}

\addinstitution{
Paderborn University \\ Paderborn, Germany
}

\runninghead{Huber et al.}{Stating comparison score uncertainty}

\begin{document}

\maketitle

\begin{abstract}
Face Recognition (FR) is increasingly used in critical verification decisions and thus, there is a need for assessing the trustworthiness of such decisions. The confidence of a decision is often based on the overall performance of the model or on the image quality. We propose to propagate model uncertainties to scores and decisions in an effort to increase the transparency of verification decisions. This work presents two contributions. First, we propose an approach to estimate the uncertainty of face comparison scores. Second, we introduce a confidence measure of the system's decision to provide insights into the verification decision. The suitability of the comparison scores uncertainties and the verification decision confidences have been experimentally proven on three face recognition models on two datasets.
\end{abstract}

\section{Introduction}

When human operators are asked if two face images show the same individual, they can intuitively state how sure they are about their decisions \cite{metacognition}. They may even conclude that they cannot make a meaningful, justifiable decision. The estimation of confidence and uncertainty regarding decision-making is important and common for humans \cite{pfe,humans}. However, state-of-the-art face recognition models (FR) \cite{magface, curricularface, DBLP:conf/cvpr/BoutrosDKK22} do not offer these confidence or uncertainty estimates, even as they outperform humans in some cases \cite{humans}. In state-of-the-art FR systems, the similarity of two faces is determined by the similarity of their feature representations generated by the model. This similarity is known as a comparison score and is used to conclude a matching decision depending on a given threshold. Intuitively, the distance of this score to the decision threshold can be interpreted as the confidence in the decision.
This similarity, or rather its deviation from the decision threshold in either direction, can be perceived as an intuitive confidence in the decision. However, this intuitive confidence measure might not be optimal in all cases, especially when the decision threshold is optimized to low false match rates.  

The importance of transparent decisions has a special significance in the field of FR. The importance of FR is growing in society and it is becoming ubiquitous and increasingly important in critical systems such as security, law enforcement, and access control \cite{survey}. In current automated FR systems, the decision of a match or non-match between two face images is commonly based on face templates extracted with deep learning networks, which makes creating a concrete notion of decision confidence rather a challenging one. This is further stressed by the fact that such verification decisions were shown to be biased towards a large set of factors \cite{bias}, and thus their lack of confidence transparency might lead to subsequent ethically compromised actions.

\begin{figure}[h]
    \centering
     \includegraphics[width=0.60\textwidth]{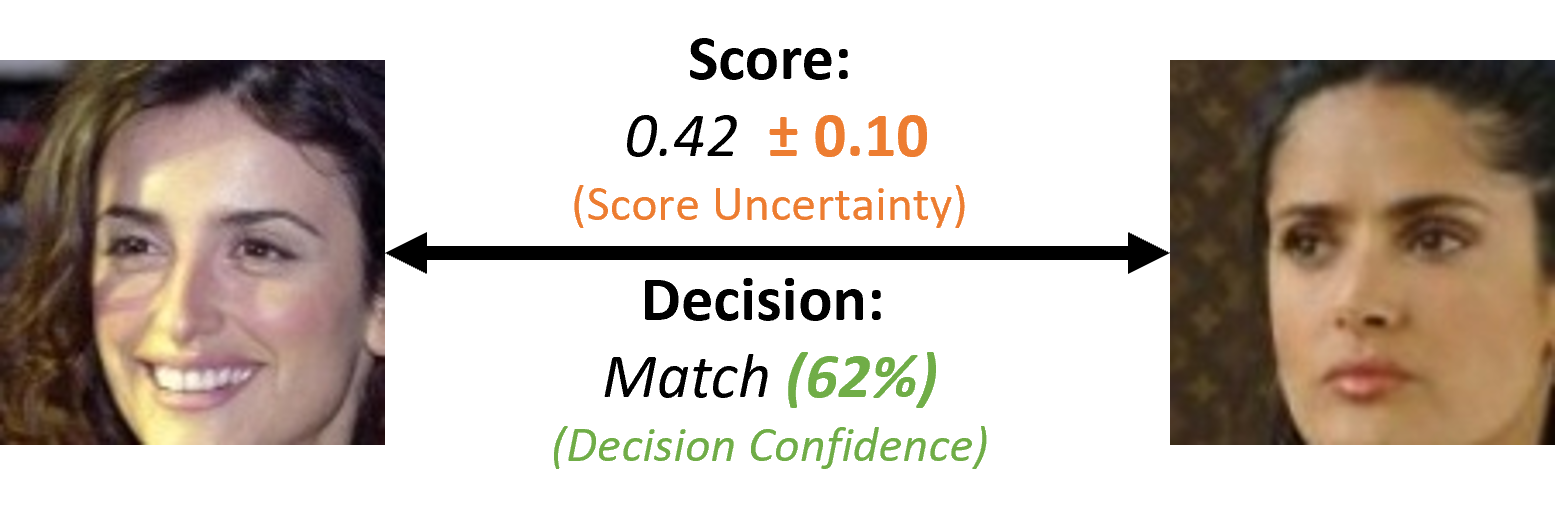}
    \caption{\textbf{Transparent Face Verification:} In addition to the usual information, such as the score and the decisions made, we propose the following to increase the transparency of the FR system: the score uncertainty (orange) and the decision confidence (green).}
    \label{fig:idea}
\end{figure}

This differentiation follows the definition of Peterson and Pitz \cite{confidence} and refers to the \textit{uncertainty} in the sense of \textit{the belief about the variability of possible outcomes} and to the \textit{confidence} in the sense of \textit{the belief that a given prediction is correct}. Both terms are therefore not interchangeable. Consequently, for face verification, we can only determine the confidence for a verification decision or the uncertainty of a comparison score. The uncertainty of a comparison score describes the uncertainty in the score depending on the uncertainty of the data and the model. In contrast, the confidence of a model's decisions describes the confidence in the made decision, with lower confidence the model is more unsure regarding its made decision. An example to illustrate the distinction between confidence and uncertainty is shown in Figure \ref{fig:idea}.

The first work that included uncertainty information in the FR process was done by Shi and Jain \cite{pfe}. They proposed probabilistic face embeddings (PFEs), that estimate the uncertainty information of a single face image by a stage-wise learning process on top of a pre-trained FR model to incorporate the uncertainty of the facial features into the embedding. Chang et al. \cite{dul} enhanced the idea of PFEs and propose to simultaneously learn the feature and the uncertainty of the facial features. Related to the determination of the uncertainty and confidence of an FR model are the works on Face Image Quality (FIQ). The idea of FIQ is to assign each face a quality estimate in the sense of the utility of a face for FR purpose \cite{fiqsurvey}. However, the interaction of two images during the comparison process is disregarded and FIQ can only be considered as an uncertainty or confidence estimate to a limited extent.

To the best of our knowledge, no work has investigated the uncertainty of a comparison score and the decision confidence in FR yet, besides interpreting the degree of similarity as the intuitive confidence. The degree of uncertainty regarding a comparison score and the statement of decision confidence deepens the transparency of the used FR model regarding its decision and might prevent incorrect decisions e.g. in front of a court.

In this work, we propose two main contributions. First, we exploit the approximation of model uncertainty through dropout and propagate this uncertainty estimate to the scoring function to obtain a score uncertainty measure. Second, we use the obtained score uncertainty to formulate a function that allows us to determine the confidence of the model regarding its decision. In experiments, we evaluate the sanity of our proposed score uncertainty and compare our proposed decision confidence with the intuitive decision confidence. The comparison score uncertainties and the uncertainty-aware decision confidence can increase the transparency without the need for any further training or adjustment.

\section{Related Work}
\label{sec:related}

In this section, we will look in more detail at PFEs that include uncertainty information into the stored face image embedding and then look at the differences and related work in the area of FIQ since no work yet focused on the exploitation of the score uncertainty. Although we take the deterministic embedding as a basis, our approach can be considered as an expression of PFEs, since we use the uncertainty of the embeddings to derive the score uncertainty. The topic of FIQ is related to the topic of FR uncertainty, as many FIQ approaches relate the utility of face images to model uncertainty.

\subsection{Probabilistic Face Embeddings}

PFEs for a single face image were first introduced by Shi and Jain \cite{pfe}. They represent each face image as Gaussian distribution with the feature as the mean and the uncertainty of the features as the variance. The variances were obtained by training a model on top of an existing FR model by utilizing the proposed mutual likelihood score. Chang et al. \cite{dul} extended the idea and proposed to train features and uncertainty simultaneously. Li et al. \cite{spherical} proposed a new optimization objective to counter some drawbacks of the approach proposed by Shi and Jain \cite{pfe}. Li et al. \cite{ordinal} proposed to learn probabilistic ordinal embeddings to be able to perform uncertainty-aware regression. Debnath et al. \cite{cluster} investigated how uncertainty in face embeddings can be used to produce face clusters of higher quality and proposed an uncertainty-aware face clustering algorithm. Chen et al. \cite{DBLP:journals/ivc/ChenYL22} proposed to improve
the robustness and speed of PFEs by simplifying the matching function and using an output-constraint loss.
None of the previous works exploited the calculated uncertainty to gain insights about the uncertainty of the scores or the confidence of the decision, which we propose in this work.

\subsection{Uncertainty and Face Image Quality}

The topic of uncertainty and FIQ are connected and not always clearly distinguishable as authors use the terms interchangeably \cite{spherical,pcnet,reid}. Outside the face recognition community, Kendall and Gal \cite{kendall} differentiate between two types of uncertainty: data uncertainty (aleatoric) that captures the noise in the input data (e.g. blurry face images) and model uncertainty (epistemic) that captures the uncertainty of the model. While FIQ assessment might be interpreted as estimating the data uncertainty independent from the model, approaches exist that combine both uncertainties to get model-specific quality estimates. SER-FIQ \cite{serfiq} utilized the Bayesian approximation of model uncertainty via dropout to calculate model-specific FIQ scores. MagFace \cite{magface} even incorporates the quality as the magnitude into the created face embeddings. QMagFace \cite{DBLP:journals/corr/abs-2111-13475} goes even further and incorporates FIQ into the comparison process. In contrast, EQFace \cite{eqface} attached a quality-prediction network on an FR model to include FIQ scores in the training process.

It is important to note that FIQ always refers to a single image and does not consider the comparison itself. FIQ values estimate the overall utility of an image to be used in FR and not for a specific comparison pair which is in contrast to our approach. Our approach takes both images, reference and probe, into account to estimate the uncertainty. Some works utilized the comparison in different ways. In PCNet \cite{pcnet}, the authors proposed to start from the confidence of mated pairs to derive an FIQ score for both involved images. Ou et al. \cite{sddfiqa} utilize the similarities between intra- and inter-class samples to derive a quality pseudo-label and therefore also incorporate both, model and data uncertainty and Bhattacharya et al. \cite{sfqa} interpreted the mean of the FIQ scores of a pair as a Face Quality Confidence Score.

\section{Methodology}
\label{sec:method}

We propose two main contributions: 1) the score uncertainty methodology that allows the interpretation of the propagated error of the comparison score, 2) a decision confidence approach that allows the interpretation of the reliability of a decision. Since both of these contributions build on the model uncertainties on the feature-level, we will first introduce how these are estimated.

\subsection{Embedding Uncertainty Estimation }
\label{subsec:uncertainty}

\begin{figure}
    \centering
    \includegraphics[width=0.7\textwidth]{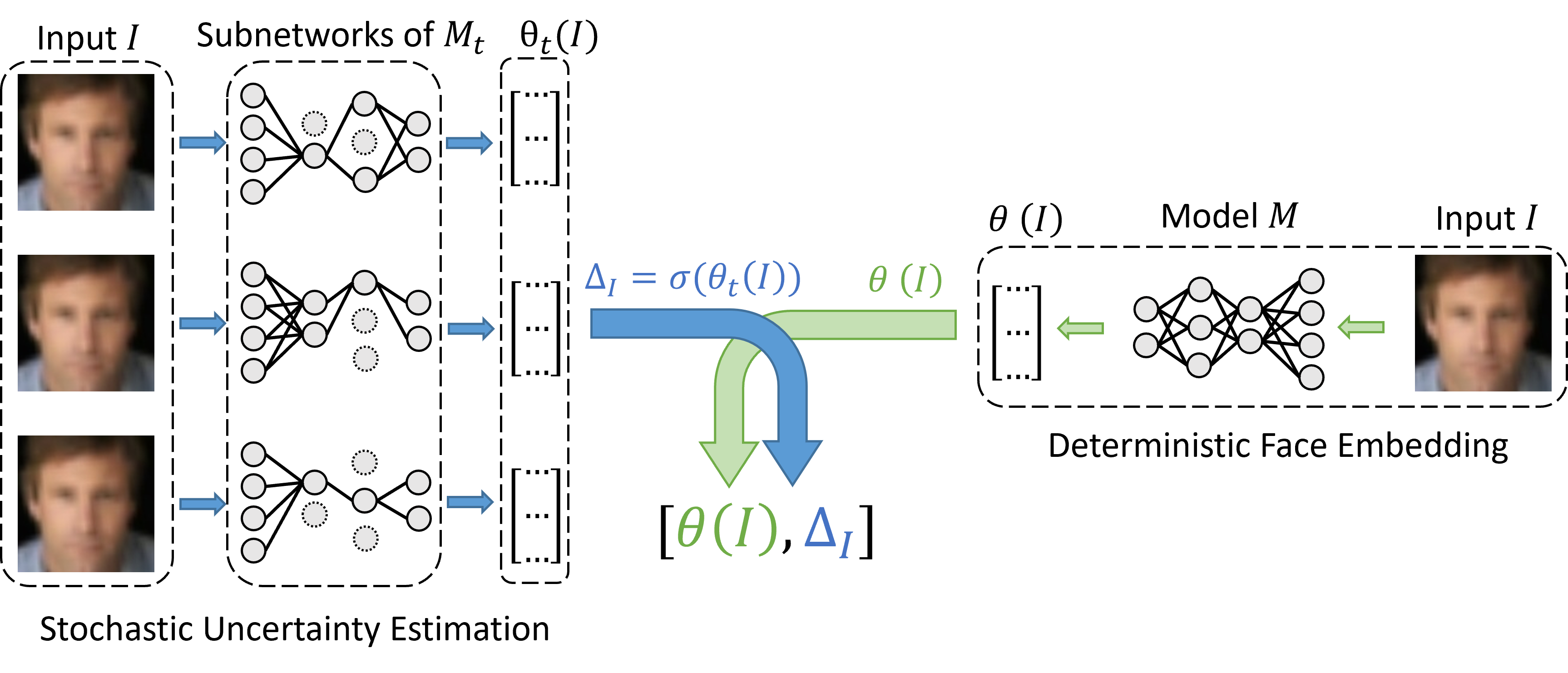}
    \caption{\textbf{Visualization of the uncertainty estimation of the embedding:} The input image $I$ is used to create the deterministic representation $\theta (I)$. $I$ is also propagated $t$ times through different subnetworks of $M$ to create a set of stochastic embeddings $\theta_{t}(I)$. These stochastic embeddings are used to estimate the uncertainty of the embedding in form of the standard deviation $\Delta_{I} = \sigma(\theta_{t}(I))$.}
    \label{fig:embedding_uncertainty}

\end{figure}

Given a face image $I$ and an FR model $M$ that provides a face embedding function $\theta$, we use $\theta$ to create a deterministic feature representation $\theta(I)$. To estimate the model uncertainty, we apply $t=100$ stochastic forward passes (as recommended by Terhörst et al. \cite{serfiq}) with different dropout patterns being applied to $M$.  This was mathematically proved by Gal and Ghahramani \cite{gal} to estimate the model's uncertainty. Doing this, we create a set of stochastic embeddings and calculate the uncertainty $\Delta_{I}$ of the feature representation for each feature dimension of the image $I$ as the standard deviation of this set. We obtain a probabilistic embedding consisting of the deterministic embedding and its uncertainty in each feature dimension. A visualization of the uncertainty estimation process of the embedding can be seen in Figure \ref{fig:embedding_uncertainty}. The computational burden can be reduced by limiting the stochastic forward passes to the last layer.

\subsection{Score Uncertainty}
\label{subsec:propagation}

After the calculation of the uncertainty $\Delta_{I}$ for an embedding $\theta(I)$ of an image $I$, we obtain a probabilistic face embedding consisting of the actual deterministic embedding and the uncertainty in each feature dimension. To decide whether two embeddings represent the same identity or not, we need a decision measure. Typically, the cosine similarity is calculated to determine the similarity of two vectors that represent two face images \cite{arcface,curricularface}, but theoretically the method can be applied to any differentiable comparison function. The cosine similarity $S_{C}$ is defined as
\begin{equation}
    S_{C}(X,Y) = \frac{X \cdot Y}{\lVert X \lVert \lVert Y \lVert}
\end{equation}
where X and Y are face embeddings. Since the embeddings are often normalized before the comparisons, and thus mapped to a unit-sphere, the formula simplifies to: 
\begin{equation}
    S_{C}(X,Y) = \sum_{n=1}^{N} x_{n} \cdot y_{n}
\end{equation}

as $\lVert X \lVert \lVert Y \lVert= 1$. $N$ denotes the number of dimensions of the embeddings $X, Y$. To obtain the uncertainty of the comparison score, we make use of the propagation of uncertainty formula, presented in \cite{uncprop}, while neglecting cross-feature correlations and assuming independence of the dimensions:
\begin{equation}
    \Delta_{E} = \sqrt{\left({\frac{\partial f}{\partial x}}\right )^2 \cdot \Delta^2}    
\end{equation}
were $\Delta_{E}$ denotes the propagated uncertainty. The assumption of independence and the neglecting of the correlations was for computational simplification. 
An investigation of the features and their correlations leads to the insight, that the dimensions are to some degree disentangled and ways to incorporate or estimate the cross-correlations are open for future research.
We set $f$ to the cosine similarity function and combine both embeddings, $X$ and $Y$ to $Z=[X,Y]$. By this, we achieve:

\begin{equation} \label{eq4}
\begin{split}
\Delta_{S_{C}} & = \sqrt{\sum_{i=1}^{2n} \left({\frac{\partial f}{\partial z_{i}}} \right )^2 \cdot \Delta_{i}^2} \\
& = \sqrt{\sum_{i=1}^n \left({\frac{\partial f}{\partial x_{i}}}\right)^2 \cdot \Delta_{x_{i}}^2} + \sqrt{\sum_{i=1}^n \left ({\frac{\partial f}{\partial y_{i}}}\right )^2 \cdot \Delta_{y_{i}}^2} \\
& = \sqrt{\sum_{i=1}^{n} y_{i}^{2} \cdot \Delta_{x_{i}}^{2} + \sum_{i=1}^{n} x_{i}^{2} \cdot \Delta_{y_{i}}^{2}}
\end{split}
\end{equation}

$\Delta_{S_{C}}$, therefore, estimates the uncertainty of the comparison score $S_{C}$ based on the individual uncertainties of   the feature of both embeddings of the comparison $X, Y$. This estimation of the score uncertainty allows a deeper insight into the calculated similarity score as it additionally takes the uncertainty of the representation into account. If the uncertainty in the face image representation is larger, it also represents a higher uncertainty in the comparison score. 

\subsection{Decision Confidence}

Since the score uncertainty $\Delta_{S_{C}}$ only indicates how certain the model is about the score and not how confident it is about a decision made, we also propose a decision confidence approach. Following the definition of confidence, which refers to a decision and the belief that the decision made is correct, we incorporate the decision threshold $d$ into our approach. Since we propagate uncertainties to derive the decision confidence, a derivable decision function is needed. Therefore, we approximate the decision function, usually in the form of a step function, by a modified sigmoid function $\delta$:
\begin{equation}
    \delta (S_{C}) = \frac{1}{1+e^{- \alpha (S_{C} - d)}}
\end{equation}
where $d$ denominates the given decision threshold, $S_{C}$ the comparison score, and $\alpha$ a scaling parameter to scale the range of the decision confidence. The $\alpha$ value represents how sharp it resembles the decision step function and thus, it weights the trade-off between the importance of the uncertainty and the distance to the decision threshold. We now also apply the propagation of uncertainty formula \cite{uncprop} to gain a confidence estimate of the decision made depending on the decision threshold and on the uncertainty of the embeddings. The confidence estimate is defined as:
\begin{equation} \label{eq6}
\begin{split}
    \Lambda_{d}(S_{C}) = 1 - \left[\sum^{N} \delta (S_{C}) \cdot (1-\delta(S_{C})) \cdot \alpha x_{n} \cdot \Delta_{y_{n}}^{2}  \right. \left. + \sum^{N} \delta (S_{C}) \cdot (1-\delta(S_{C})) \cdot \alpha y_{n} \cdot \Delta_{x_{n}}^{2} \right ] ^{\frac{1}{2}} \\
\end{split}
\end{equation}
where $x_{n}$ and $y_{n}$ refers to features of the embeddings $X$ and $Y$ and $\Delta_{x_{n}}$, and $\Delta_{y_{n}}$, to the corresponding feature uncertainties. To obtain the confidence, we subtract the propagated uncertainty from 1 to get the intuitive understanding from increasing and decreasing confidence in a natural sense.

\section{Experimental Setup}
\label{sec:experiments}

\subsection{Face Recognition Models}
\label{subsec: frmodels}

To show the validity of our proposed uncertainty scores and decisions confidences, we use three different well-performing FR models as a basis. All models share the same ResNet-100 \cite{resnet} architecture and have been trained with their specific loss function. The models were trained on the MS1M-V2 \cite{arcface} dataset and were officially provided as pre-trained models by the authors. The used models are named ArcFace \cite{arcface}, CurricularFace \cite{curricularface}, and MagFace \cite{magface}. We use these models to show that our proposed uncertainties and confidences are model-independent and can be transferred to a range of diverse models.

\subsection{Datasets \& Evaluation Benchmarks}
\label{subsec:benchmarks}

To investigate the meaningfulness of our proposed uncertainty estimation approach and the proposed decision confidence, we evaluate on two difficult datasets with a higher variance in the quality of the images. We chose the Adience dataset \cite{adience} that consists of 26k images of over 2k different subjects as the images vary in a wide range of image quality, age, and pose. For the same reason, we also chose the XQLFW \cite{xqlfw} dataset that is based on the LFW \cite{lfw} dataset and consists of synthetically degraded images as well as unaltered ones. Originally, intended to evaluate performance on cross-quality images, we use it to show that our uncertainty can also handle low-quality pairs as well as high-quality pairs. For this, we do not limit ourselves to the defined pairs for the evaluation of the uncertainty and the confidence and consider additional pairs. Both datasets are used to evaluate the score uncertainty and the decision confidence. 

\subsection{Evaluation Metrics}
\label{metrics}

To investigate the rationale of our estimated uncertainty and our decision confidence, we use Error-vs-Reject curves (ERC) \cite{evr}. ERCs were proposed as a measure for the performance of a biometric quality estimator and they show the verification error rate over the fraction of unconsidered face images, where the images with the lowest quality are removed. Starting from the whole set of pairs, we subsequently remove the pairs with the highest uncertainty/lowest confidence and evaluate the performance again. We adapt the ERC and use them to evaluate the expressiveness of our uncertainty and confidence. For the uncertainty, the accuracy should ideally increase if uncertain scores are removed. For the confidence, the accuracy should increase, if pairs with low decision confidence are removed. For the experiments, we take all genuine pairs and select 30 random imposter images for each image from all possible imposter images.
For the investigation of the usefulness of the score uncertainty, we compare our proposed score uncertainty with the MagFace \cite{magface} FIQ values. Although FIQ is designed and intended to be used to estimate the quality of single images, some evaluation protocols do suggest rejecting the comparison pairs if one of the images in the pair is of low-quality \cite{magface}. Based on that, we provide a minimized evaluation to compare the performance of such a technique (despite its limitations). On all ERC, we report the verification performance in terms of false non-match rate (FNMR) at fixed false match rate (FMR) following the international standard \cite{ISO_Metrik}. We also follow the best practice for high-security scenarios, e.g. automatic border control systems proposed by the European Border and Coast Guard Agency (Frontex) \cite{frontex2015best} and evaluate at the operation point of fixed FMR $<$ 0.1\%. 

\begin{figure*}[h]
    \centering
     \includegraphics[width=0.80\textwidth]{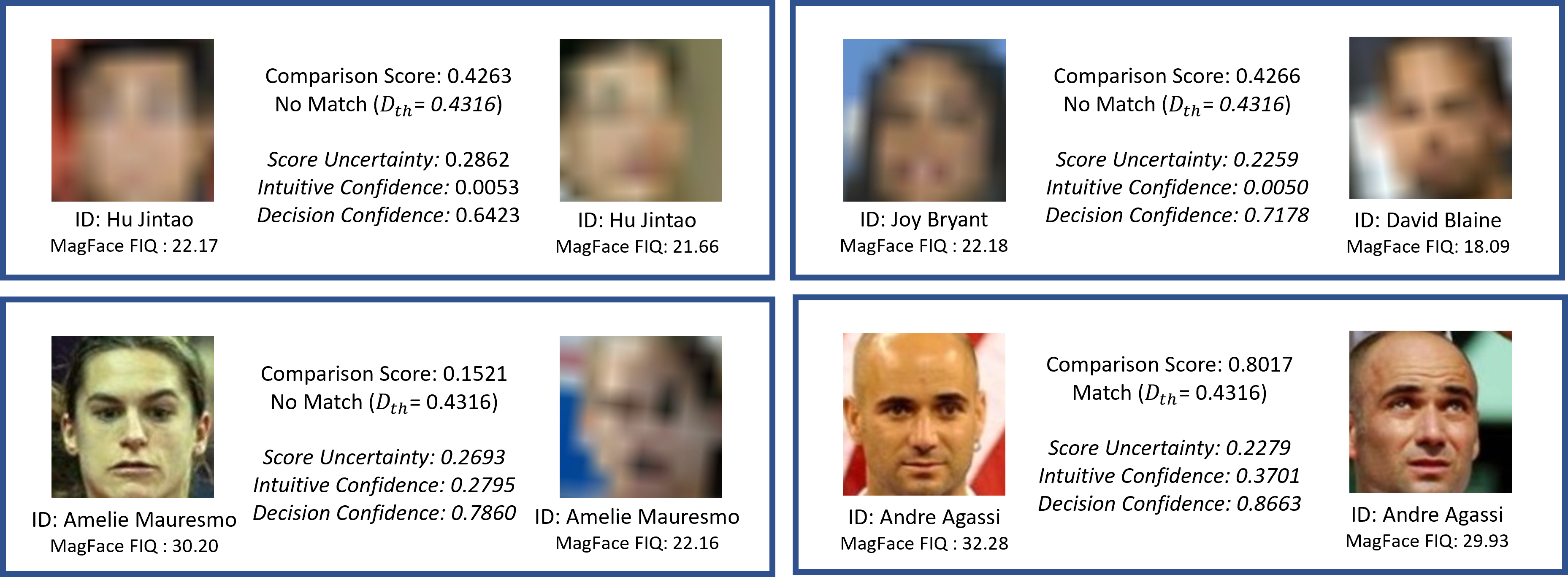}
    \caption{\textbf{Transparent Face Verification:} For each pair not only the decision is reported, but also the score uncertainty and the decision confidence. In the upper row even though the intuitive confidence is almost the same, the proposed decision confidence shows that the confidence in the false no-match decision on the left is lower in comparison to the correct no-match decision on the right. In the bottom row, the decision confidence is higher if the score uncertainty is smaller and the images are less similar.}
    \label{fig:overiew}
    
\end{figure*}

For the investigation of our decision confidence, we compare our proposed decision confidence with an intuitive confidence. This intuitive confidence is derived from the comparison score and its distance to the decision threshold $D_{th}$. The decision threshold $D_{th}$ is also determined at the fixed FMR and does not change as comparisons are removed. The intuitive idea is, that the confidence of a correct decision increases with a higher or lower similarity, starting from the decision threshold that divides the similarity range into match and no match. The intuitive confidence $\Lambda_{Int}$ can therefore be defined as:
\begin{equation}
    \Lambda_{Int}(s) = |s - D_{th}|
\end{equation}
where $s$ is the given similarity score and $D_{th}$ is the decision threshold of the system. A visualization of the intuitive confidence is shown in Figure \ref{intconf}. It is important to keep in mind, that as the intuitive confidence only depends on the similarity score and its distance to the threshold all pairs with the same similarity score share the same confidence.

\section{Results}
\label{sec:results}

In this section, we investigate the performance of our two contributions separately. We start with evaluating the estimated score uncertainty by looking at ERCs to investigate the sanity of the proposed approach. In a second evaluation, we compare our proposed decision confidence against the intuitive confidence on two datasets to show that a deeper interpretation of the system's confidence in a decision is possible. An overview of all contributions combined can be seen in Figure \ref{fig:overiew}.

\subsection{Investigating Score Uncertainty}

\begin{figure}
    \centering
    \begin{subfigure}[Adience]{ %
     \includegraphics[width=0.40\textwidth]{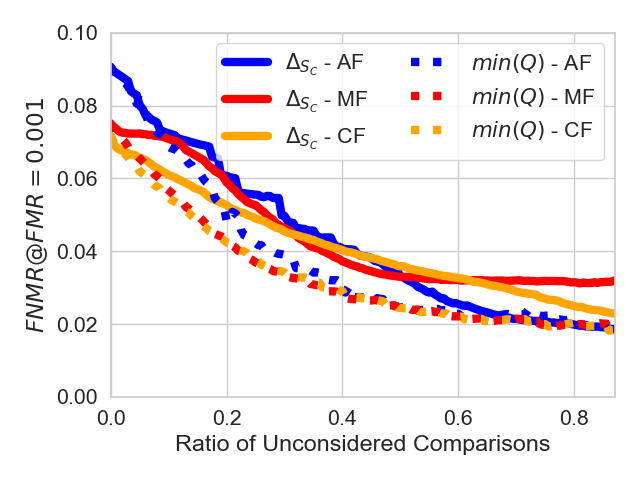}}
     \end{subfigure}
     \begin{subfigure}[XQLFW]{ %
     \includegraphics[width=0.40\textwidth]{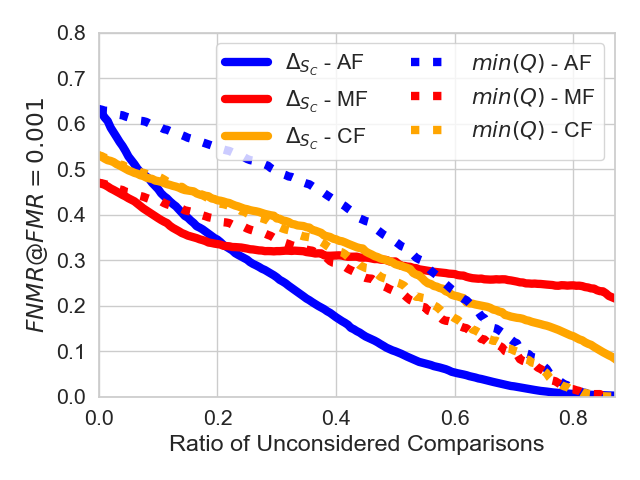}}
     \end{subfigure}
    \caption{\textbf{Score uncertainty evaluation using ERC:} The comparisons are unconsidered either based on their score uncertainty or on their minimum quality based on the MagFace \cite{magface} embedding quality score. With a higher uncertainty regarding the score, more wrong decisions arise, accordingly the error decreases when the comparisons with the highest uncertainty are removed. The steadily decrease indicating the strong usefulness of the estimated uncertainties. (AF = ArcFace \cite{arcface}, MF = MagFace \cite{magface}, CF = CurricularFace \cite{curricularface})}
    \label{fig:unc_erc}
    
\end{figure}

To evaluate the utility of our proposed score uncertainty we plot ERCs on two datasets, Adience \cite{adience} and XQLFW \cite{xqlfw}, at the operation point of FMR=0.001 and report the performance as the FNMR at this point. We compare our proposed score uncertainty in a minimized evaluation with the FIQ provided by the MagFace embeddings. The ERCs are shown in Figure \ref{fig:unc_erc}. We can observe that performance improves, i.e. fewer errors are made, when those comparisons are removed where the score has the highest uncertainty. This can be observed on all three models on both utilized datasets. Even if the minimum quality is in some cases better than the proposed uncertainty, we emphasize the differences: while the minimum quality approach reduces the problem of the inaccuracy of the comparison to the lowest quality, our approach processes the inaccuracy of the mapping and considers both images.

\subsection{Analysing Decision Confidence}

\begin{figure*}
    \centering
    \begin{subfigure}[AF - Adience]{ 
     \includegraphics[width=0.3\textwidth]{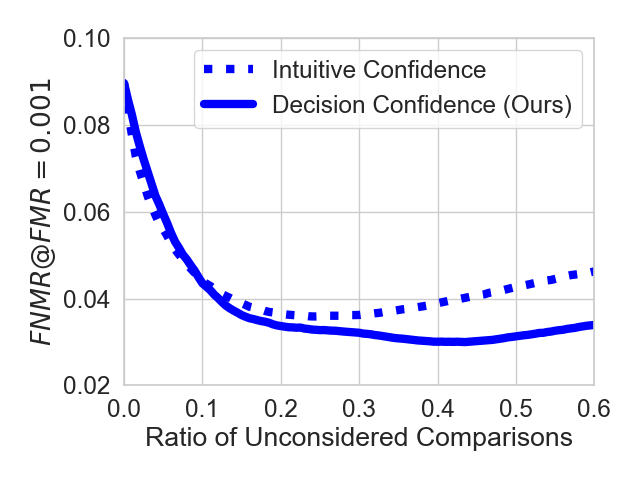}}
    \end{subfigure}
    \begin{subfigure}[MF - Adience]{ 
     \includegraphics[clip, width=0.3\textwidth]{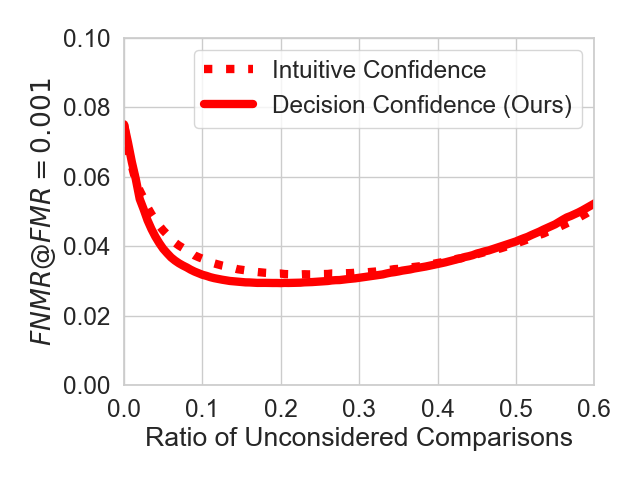}}
    \end{subfigure}
    \begin{subfigure}[CF - Adience]{ 
     \includegraphics[clip, width=0.3\textwidth]{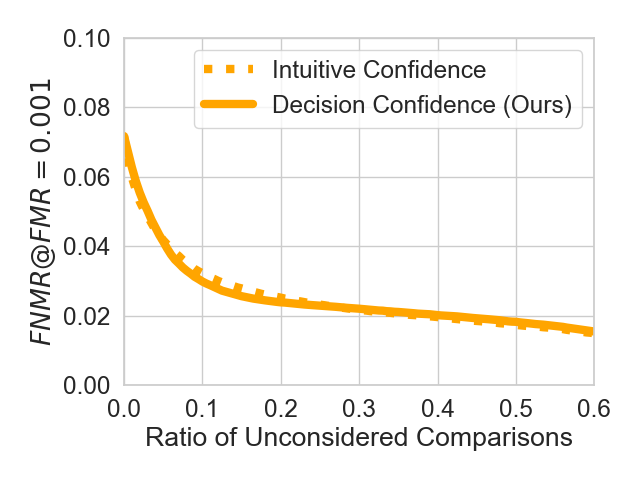}}
    \end{subfigure}
    
    \begin{subfigure}[AF - XQLFW]{
     \includegraphics[clip, width=0.3\textwidth]{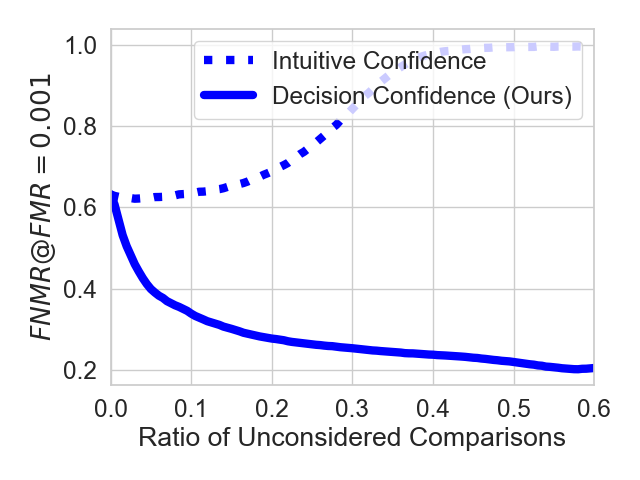}}
    \end{subfigure}
    \begin{subfigure}[MF - XQLFW]{ 
     \includegraphics[clip, width=0.3\textwidth]{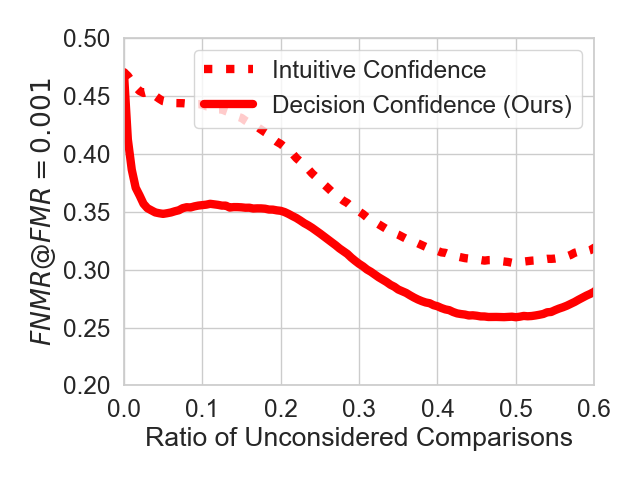}}
    \end{subfigure}
    \begin{subfigure}[CF - XQLFW]{ %
     \includegraphics[clip, width=0.3\textwidth]{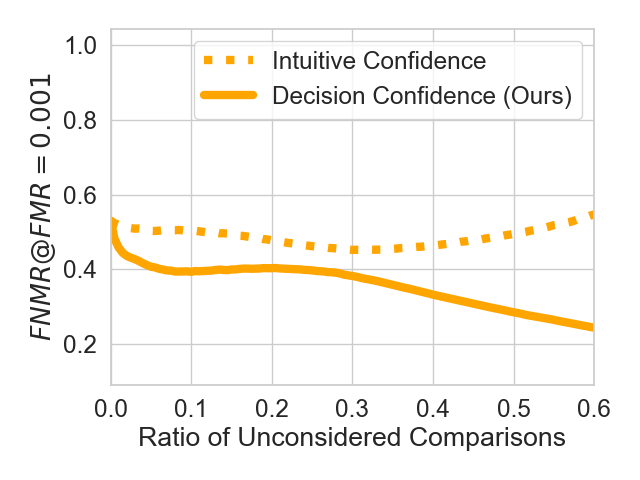}}
    \end{subfigure}
    \caption{\textbf{Evaluating Decision Confidence (FMR@0.001):} The intuitive and the proposed confidence are compared. Overall the proposed decision confidence estimated the reliability of the decision more accurate than the intuitive decision confidence.}
    \label{fig:conf_erc}
\end{figure*}

\begin{figure}
    \centering
    \begin{subfigure}[Intuitive Confidence\label{intconf}]{ %
     \includegraphics[width=0.35\textwidth]{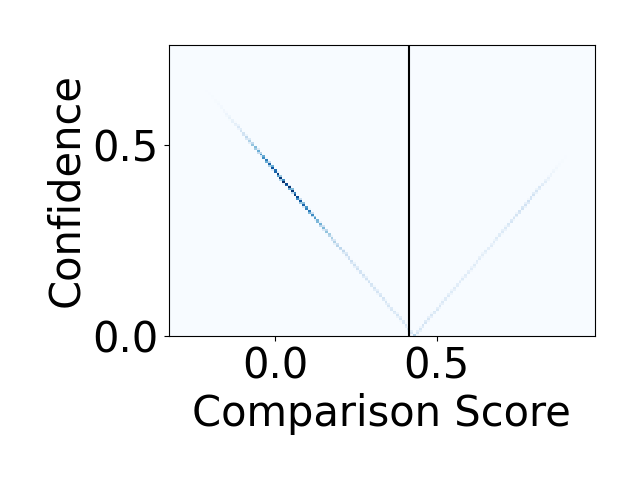}}
    \end{subfigure}
    \begin{subfigure}[Decision Confidence\label{conf}]{ %
     \includegraphics[width=0.35\textwidth]{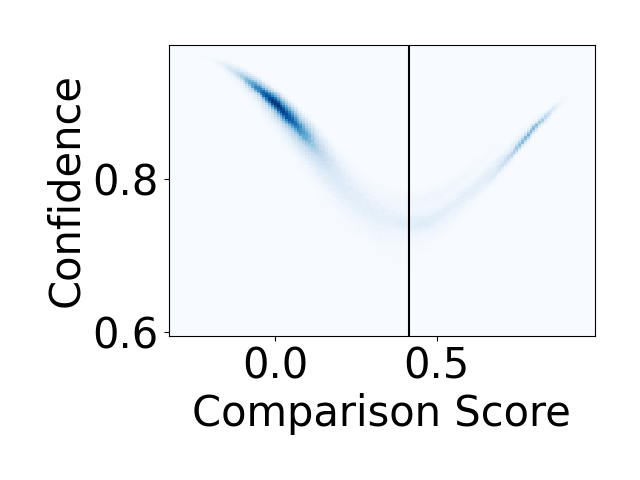}}
    \end{subfigure}
    \caption{\textbf{Comparison of the confidence measures (Heatmap):} On the Adience dataset using the MagFace model. While the intuitive confidence only depends on the comparison score and its distance to the decision threshold (black asymptote), the proposed decision confidence also considers the uncertainty of the score and provides a more natural understanding of confidence.}
    \label{fig:heat}
    \vspace{-5mm}
\end{figure}

For the evaluation of the decision confidence, we also investigate the performance by looking at ERCs and compare our proposed decision confidence with the intuitive decision confidence as defined above in Subsection \ref{metrics}. It is important to note that the decision confidence is threshold dependent as it defines the confidence about a made decision. The results are presented in Figure \ref{fig:conf_erc} on the operational point of 0.1\% FMR. For $\alpha$, we chose for the ArcFace model $\alpha=2$ and for the other models, MagFace and CurricularFace $\alpha=5$. Other values for $\alpha$ are shown in the supplementary material. With the $\alpha$-parameter the influence of the threshold can be scaled in relation to the uncertainty. On the Adience dataset, we observe that a minor improvement can be achieved using the proposed decision confidence in comparison to the intuitive confidence. Therefore, it is possible and useful to find more sophisticated ways of interpreting the confidence of a model's decision besides the similarity of pairs and the distance to an operational threshold. A larger improvement can be observed in the experiments on the harder XQLFW dataset. As we do not limit the pairs to be cross-quality but also take high- and low-quality pairs into consideration, the decision confidence shows a higher increase in reliability of the made decision. 
For a deeper insight into our proposed decision confidence in comparison to the intuitive confidence, we show in Figure \ref{fig:heat} the heatmaps of the confidences depending on the score. While for the intuitive solution the same comparison score leads to the same degree of confidence, in our proposed approach the uncertainty also is taken into consideration. This allows a more natural interpretation of confidence as it not only depends on the similarity but also on the uncertainty of the similarity. 

\section{Conclusion}
\label{sec:conclusion}

In this work, we proposed 1) to estimate the uncertainty of the comparison score of FR models, 2) to provide decision confidence on the FR systems' decision. After showing the meaningfulness of our proposed score uncertainty, we proposed to use the score uncertainty to derive decision confidence that is able to beat the intuitive confidence derived from the similarity of the embedded face images and the distance to the decision threshold. By introducing the new concepts of score uncertainty and decision confidence we highlight the need for transparency in the biometric decision-making process. This is necessary to increase the trust and reliability of FR systems and to draw attention to possible biases in form of lacking decision confidence, especially as FR becomes more ubiquitous and widespread in our society.

\subsubsection*{Acknowledgement}

This research work has been funded by the German Federal Ministry of Education and Research and the Hessian Ministry of Higher Education, Research, Science and the Arts within their joint support of the National Research Center for Applied Cybersecurity ATHENE. This work was also partially carried out during the tenure of an ERCIM ’Alain Bensoussan‘ Fellowship Programme.




\bibliography{egbib}

\begin{thebibliography}{33}
\providecommand{\natexlab}[1]{#1}
\providecommand{\url}[1]{\texttt{#1}}
\expandafter\ifx\csname urlstyle\endcsname\relax
  \providecommand{\doi}[1]{doi: #1}\else
  \providecommand{\doi}{doi: \begingroup \urlstyle{rm}\Url}\fi

\bibitem[Bhattacharya et~al.(2021)Bhattacharya, Kyal, and Routray]{sfqa}
Shubhobrata Bhattacharya, Chirag Kyal, and Aurobinda Routray.
\newblock Simplified face quality assessment {(SFQA)}.
\newblock \emph{Pattern Recognit. Lett.}, 147:\penalty0 108--114, 2021.

\bibitem[Boutros et~al.(2022)Boutros, Damer, Kirchbuchner, and
  Kuijper]{DBLP:conf/cvpr/BoutrosDKK22}
Fadi Boutros, Naser Damer, Florian Kirchbuchner, and Arjan Kuijper.
\newblock Elasticface: Elastic margin loss for deep face recognition.
\newblock In \emph{{CVPR} Workshops}, pages 1577--1586. {IEEE}, 2022.

\bibitem[Chang et~al.(2020)Chang, Lan, Cheng, and Wei]{dul}
Jie Chang, Zhonghao Lan, Changmao Cheng, and Yichen Wei.
\newblock Data uncertainty learning in face recognition.
\newblock In \emph{{CVPR}}, pages 5709--5718. Computer Vision Foundation /
  {IEEE}, 2020.

\bibitem[Chen et~al.(2022)Chen, Yi, and Lv]{DBLP:journals/ivc/ChenYL22}
Kai Chen, Taihe Yi, and Qi~Lv.
\newblock Fast and reliable probabilistic face embeddings based on constrained
  data uncertainty estimation.
\newblock \emph{Image Vis. Comput.}, 121:\penalty0 104429, 2022.

\bibitem[Debnath et~al.(2021)Debnath, Coviello, Yang, and Chakradhar]{cluster}
Biplob Debnath, Giuseppe Coviello, Yi~Yang, and Srimat Chakradhar.
\newblock Uac: An uncertainty-aware face clustering algorithm.
\newblock In \emph{Proceedings of the IEEE/CVF International Conference on
  Computer Vision (ICCV) Workshops}, pages 3487--3495, October 2021.

\bibitem[Deng et~al.(2019)Deng, Guo, Xue, and Zafeiriou]{arcface}
Jiankang Deng, Jia Guo, Niannan Xue, and Stefanos Zafeiriou.
\newblock Arcface: Additive angular margin loss for deep face recognition.
\newblock In \emph{{CVPR}}, pages 4690--4699. Computer Vision Foundation /
  {IEEE}, 2019.

\bibitem[Eidinger et~al.(2014)Eidinger, Enbar, and Hassner]{adience}
Eran Eidinger, Roee Enbar, and Tal Hassner.
\newblock Age and gender estimation of unfiltered faces.
\newblock \emph{{IEEE} Trans. Inf. Forensics Secur.}, 9\penalty0 (12):\penalty0
  2170--2179, 2014.

\bibitem[Frontex(2015)]{frontex2015best}
Frontex.
\newblock Best practice technical guidelines for automated border control (abc)
  systems, 2015.

\bibitem[Gal and Ghahramani(2016)]{gal}
Yarin Gal and Zoubin Ghahramani.
\newblock Dropout as a bayesian approximation: Representing model uncertainty
  in deep learning.
\newblock In \emph{{ICML}}, volume~48 of \emph{{JMLR} Workshop and Conference
  Proceedings}, pages 1050--1059. JMLR.org, 2016.

\bibitem[Grother and Tabassi(2007)]{evr}
Patrick Grother and Elham Tabassi.
\newblock Performance of biometric quality measures.
\newblock \emph{{IEEE} Trans. Pattern Anal. Mach. Intell.}, 29\penalty0
  (4):\penalty0 531--543, 2007.

\bibitem[He et~al.(2016)He, Zhang, Ren, and Sun]{resnet}
Kaiming He, Xiangyu Zhang, Shaoqing Ren, and Jian Sun.
\newblock Deep residual learning for image recognition.
\newblock In \emph{{CVPR}}, pages 770--778. {IEEE} Computer Society, 2016.

\bibitem[Huang et~al.(2007)Huang, Ramesh, Berg, and Learned-Miller]{lfw}
Gary~B. Huang, Manu Ramesh, Tamara Berg, and Erik Learned-Miller.
\newblock Labeled faces in the wild: A database for studying face recognition
  in unconstrained environments.
\newblock Technical Report 07-49, University of Massachusetts, Amherst, October
  2007.

\bibitem[Huang et~al.(2020)Huang, Wang, Tai, Liu, Shen, Li, Li, and
  Huang]{curricularface}
Yuge Huang, Yuhan Wang, Ying Tai, Xiaoming Liu, Pengcheng Shen, Shaoxin Li,
  Jilin Li, and Feiyue Huang.
\newblock Curricularface: Adaptive curriculum learning loss for deep face
  recognition.
\newblock In \emph{{CVPR}}, pages 5900--5909. Computer Vision Foundation /
  {IEEE}, 2020.

\bibitem[ISO/IEC 19795-1:2006)()]{ISO_Metrik}
ISO/IEC 19795-1:2006).
\newblock {ISO/IEC 19795-1:2006 Information technology — Biometric
  performance testing and reporting}.
\newblock Standard, International Organization for Standardization, 2016.

\bibitem[Kendall and Gal(2017)]{kendall}
Alex Kendall and Yarin Gal.
\newblock What uncertainties do we need in bayesian deep learning for computer
  vision?
\newblock In \emph{{NIPS}}, pages 5574--5584, 2017.

\bibitem[Knoche et~al.(2021)Knoche, H{\"{o}}rmann, and Rigoll]{xqlfw}
Martin Knoche, Stefan H{\"{o}}rmann, and Gerhard Rigoll.
\newblock Cross-quality {LFW:} {A} database for analyzing cross-resolution
  image face recognition in unconstrained environments.
\newblock \emph{CoRR}, abs/2108.10290, 2021.

\bibitem[Ku(1966)]{uncprop}
H.H. Ku.
\newblock Notes on the use of propagation of error formulas.
\newblock \emph{Journal of Research of the National Bureau of Standards.
  Section C: Engineering and Instrumentation}, 70C, 1966.

\bibitem[Li et~al.(2021{\natexlab{a}})Li, Xu, Xu, Shen, Li, and
  Hooi]{spherical}
Shen Li, Jianqing Xu, Xiaqing Xu, Pengcheng Shen, Shaoxin Li, and Bryan Hooi.
\newblock Spherical confidence learning for face recognition.
\newblock In \emph{{CVPR}}, pages 15629--15637. Computer Vision Foundation /
  {IEEE}, 2021{\natexlab{a}}.

\bibitem[Li et~al.(2021{\natexlab{b}})Li, Huang, Lu, Feng, and Zhou]{ordinal}
Wanhua Li, Xiaoke Huang, Jiwen Lu, Jianjiang Feng, and Jie Zhou.
\newblock Learning probabilistic ordinal embeddings for uncertainty-aware
  regression.
\newblock In \emph{{CVPR}}, pages 13896--13905. Computer Vision Foundation /
  {IEEE}, 2021{\natexlab{b}}.

\bibitem[Liu and Tan(2021)]{eqface}
Rushuai Liu and Weijun Tan.
\newblock Eqface: {A} simple explicit quality network for face recognition.
\newblock In \emph{{CVPR} Workshops}, pages 1482--1490. Computer Vision
  Foundation / {IEEE}, 2021.

\bibitem[Meng et~al.(2021)Meng, Zhao, Huang, and Zhou]{magface}
Qiang Meng, Shichao Zhao, Zhida Huang, and Feng Zhou.
\newblock Magface: {A} universal representation for face recognition and
  quality assessment.
\newblock In \emph{{CVPR}}, pages 14225--14234. Computer Vision Foundation /
  {IEEE}, 2021.

\bibitem[Ou et~al.(2021)Ou, Chen, Zhang, Huang, Li, Li, Li, Cao, and
  Wang]{sddfiqa}
Fu{-}Zhao Ou, Xingyu Chen, Ruixin Zhang, Yuge Huang, Shaoxin Li, Jilin Li, Yong
  Li, Liujuan Cao, and Yuan{-}Gen Wang.
\newblock {SDD-FIQA:} unsupervised face image quality assessment with
  similarity distribution distance.
\newblock In \emph{{CVPR}}, pages 7670--7679. Computer Vision Foundation /
  {IEEE}, 2021.

\bibitem[Peterson and Pitz(1988)]{confidence}
Dane~K Peterson and Gordon~F Pitz.
\newblock Confidence, uncertainty, and the use of information.
\newblock \emph{Journal of Experimental Psychology: Learning, Memory, and
  Cognition}, 14\penalty0 (1):\penalty0 85, 1988.

\bibitem[Phillips and O'Toole(2014)]{humans}
P.~Jonathon Phillips and Alice~J. O'Toole.
\newblock Comparison of human and computer performance across face recognition
  experiments.
\newblock \emph{Image Vis. Comput.}, 32\penalty0 (1):\penalty0 74--85, 2014.

\bibitem[Schlett et~al.(2020)Schlett, Rathgeb, Henniger, Galbally,
  Fi{\'{e}}rrez, and Busch]{fiqsurvey}
Torsten Schlett, Christian Rathgeb, Olaf Henniger, Javier Galbally, Julian
  Fi{\'{e}}rrez, and Christoph Busch.
\newblock Face image quality assessment: {A} literature survey.
\newblock \emph{CoRR}, abs/2009.01103, 2020.

\bibitem[Shi and Jain(2019)]{pfe}
Yichun Shi and Anil~K. Jain.
\newblock Probabilistic face embeddings.
\newblock In \emph{{ICCV}}. {IEEE}, 2019.

\bibitem[Terh{\"{o}}rst et~al.(2020)Terh{\"{o}}rst, Kolf, Damer, Kirchbuchner,
  and Kuijper]{serfiq}
Philipp Terh{\"{o}}rst, Jan~Niklas Kolf, Naser Damer, Florian Kirchbuchner, and
  Arjan Kuijper.
\newblock {SER-FIQ:} unsupervised estimation of face image quality based on
  stochastic embedding robustness.
\newblock In \emph{{CVPR}}, pages 5650--5659. Computer Vision Foundation /
  {IEEE}, 2020.

\bibitem[Terh{\"{o}}rst et~al.(2021)Terh{\"{o}}rst, Ihlefeld, Huber, Damer,
  Kirchbuchner, Raja, and Kuijper]{DBLP:journals/corr/abs-2111-13475}
Philipp Terh{\"{o}}rst, Malte Ihlefeld, Marco Huber, Naser Damer, Florian
  Kirchbuchner, Kiran~B. Raja, and Arjan Kuijper.
\newblock Qmagface: Simple and accurate quality-aware face recognition.
\newblock \emph{CoRR}, abs/2111.13475, 2021.

\bibitem[Terh{\"o}rst et~al.(2021)Terh{\"o}rst, Kolf, Huber, Kirchbuchner,
  Damer, Moreno, Fierrez, and Kuijper]{bias}
Philipp Terh{\"o}rst, Jan~Niklas Kolf, Marco Huber, Florian Kirchbuchner, Naser
  Damer, Aythami~Morales Moreno, Julian Fierrez, and Arjan Kuijper.
\newblock A comprehensive study on face recognition biases beyond demographics.
\newblock \emph{IEEE Transactions on Technology and Society}, 3\penalty0
  (1):\penalty0 16--30, 2021.

\bibitem[Wang and Deng(2021)]{survey}
Mei Wang and Weihong Deng.
\newblock Deep face recognition: {A} survey.
\newblock \emph{Neurocomputing}, 429:\penalty0 215--244, 2021.

\bibitem[Xie et~al.(2020)Xie, Byrne, and Zisserman]{pcnet}
Weidi Xie, Jeffrey Byrne, and Andrew Zisserman.
\newblock Inducing predictive uncertainty estimation for face verification.
\newblock In \emph{{BMVC}}. {BMVA} Press, 2020.

\bibitem[Yeung and Summerfield(2012)]{metacognition}
Nick Yeung and Christopher Summerfield.
\newblock Metacognition in human decision-making: confidence and error
  monitoring.
\newblock \emph{Philosophical Transactions of the Royal Society B: Biological
  Sciences}, 367\penalty0 (1594):\penalty0 1310--1321, 2012.

\bibitem[Zheng et~al.(2021)Zheng, Lan, Zeng, Zhang, and Zha]{reid}
Kecheng Zheng, Cuiling Lan, Wenjun Zeng, Zhizheng Zhang, and Zheng{-}Jun Zha.
\newblock Exploiting sample uncertainty for domain adaptive person
  re-identification.
\newblock In \emph{{AAAI}}, pages 3538--3546. {AAAI} Press, 2021.

\end{thebibliography}


\begin{thebibliography}{3}
\providecommand{\natexlab}[1]{#1}
\providecommand{\url}[1]{\texttt{#1}}
\expandafter\ifx\csname urlstyle\endcsname\relax
  \providecommand{\doi}[1]{doi: #1}\else
  \providecommand{\doi}{doi: \begingroup \urlstyle{rm}\Url}\fi

\bibitem[Deng et~al.(2019)Deng, Guo, Xue, and Zafeiriou]{arcface}
Jiankang Deng, Jia Guo, Niannan Xue, and Stefanos Zafeiriou.
\newblock Arcface: Additive angular margin loss for deep face recognition.
\newblock In \emph{{CVPR}}, pages 4690--4699. Computer Vision Foundation /
  {IEEE}, 2019.

\bibitem[Huang et~al.(2020)Huang, Wang, Tai, Liu, Shen, Li, Li, and
  Huang]{curricularface}
Yuge Huang, Yuhan Wang, Ying Tai, Xiaoming Liu, Pengcheng Shen, Shaoxin Li,
  Jilin Li, and Feiyue Huang.
\newblock Curricularface: Adaptive curriculum learning loss for deep face
  recognition.
\newblock In \emph{{CVPR}}, pages 5900--5909. Computer Vision Foundation /
  {IEEE}, 2020.

\bibitem[Meng et~al.(2021)Meng, Zhao, Huang, and Zhou]{magface}
Qiang Meng, Shichao Zhao, Zhida Huang, and Feng Zhou.
\newblock Magface: {A} universal representation for face recognition and
  quality assessment.
\newblock In \emph{{CVPR}}, pages 14225--14234. Computer Vision Foundation /
  {IEEE}, 2021.

\end{thebibliography}
\end{document}


\maketitle

\section{Supplementary Materials}
This is the supplementary material to the paper \textit{Stating Comparison Score Uncertainty and Verification Decision Confidence Towards Transparent Face Recognition}. More details about the pre-trained models and their licences are provided in Section \ref{c}. In Section \ref{a} we further investigate the influence of the $\alpha$ parameter that influences the trade-off between uncertainty and intuitive confidence for the proposed decision confidence. 

\section{Information on the Used Models}
\label{c}
This section provides additional information about the used models during the experiments to increase the reproducability of the experiments. 

The ArcFace \cite{arcface} model was trained by the authors and provided at their official GitHub repository: \url{https://github.com/deepinsight/insightface}. The model can be downloaded at: \url{https://github.com/deepinsight/insightface/tree/master/model_zoo}. We utilized the R100 model trained on the MS1MV2 dataset. All models are available for non-commercial research purposes only and the code is released under MIT License.

The MagFace \cite{magface} model was trained by the authors and provided at their official GitHub repository: \url{https://github.com/IrvingMeng/MagFace}. We used the model with the iResNet100 architecture and trained on the MS1MV2 dataset. The code is released under the Apache-2.0 license.

The CurricularFace \cite{curricularface} model was trained by the authors and provided at their official GitHub repository: \url{https://github.com/HuangYG123/CurricularFace}. The code is released under the MIT License.

\section{Influence of $\alpha$ on the Decision Confidence}
\label{a}
\begin{figure*}
    \centering
    \begin{subfigure}[AF, $\alpha = 1$\label{intconf}]{ %
     \includegraphics[clip, width=0.18\textwidth]{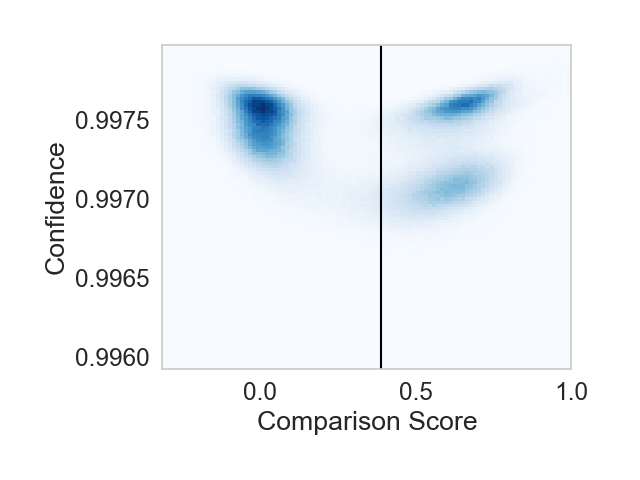}}
     \end{subfigure}
     \begin{subfigure}[AF, $\alpha = 2$\label{intconf}]{ %
     \includegraphics[clip, width=0.18\textwidth]{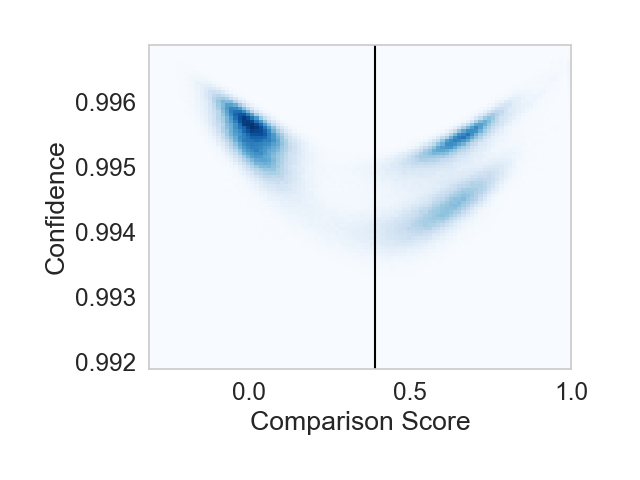}}
     \end{subfigure}
     \begin{subfigure}[AF, $\alpha = 3$\label{intconf}]{ %
     \includegraphics[clip, width=0.18\textwidth]{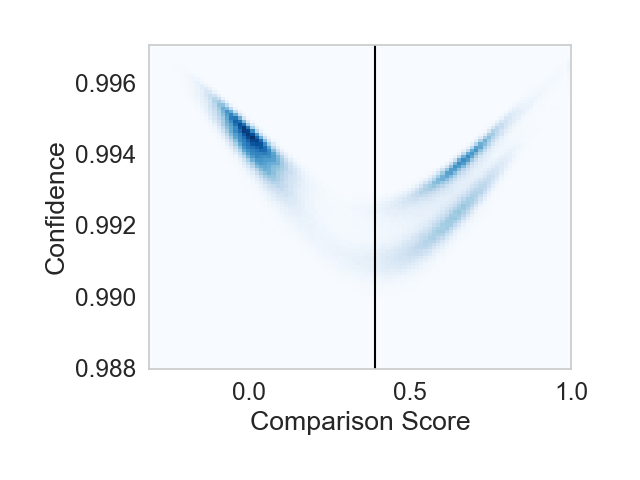}}
     \end{subfigure}
     \begin{subfigure}[AF, $\alpha = 5$\label{intconf}]{ %
     \includegraphics[clip, width=0.18\textwidth]{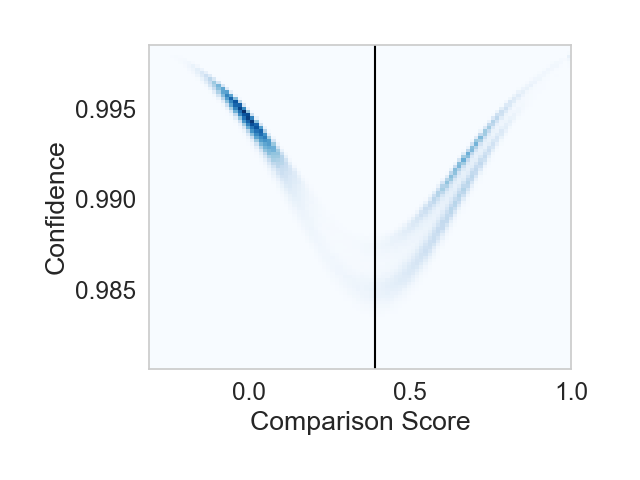}}
     \end{subfigure}
     \begin{subfigure}[AF, $\alpha = 7$\label{intconf}]{ %
     \includegraphics[clip, width=0.18\textwidth]{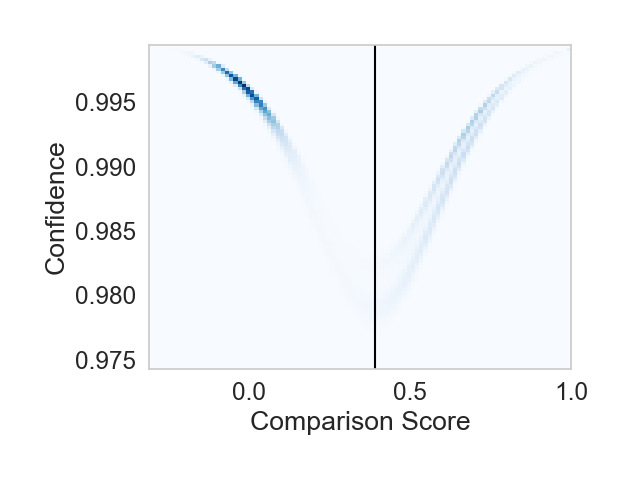}}
     \end{subfigure}
     
     \begin{subfigure}[MF, $\alpha = 1$\label{intconf}]{ %
     \includegraphics[clip, width=0.18\textwidth]{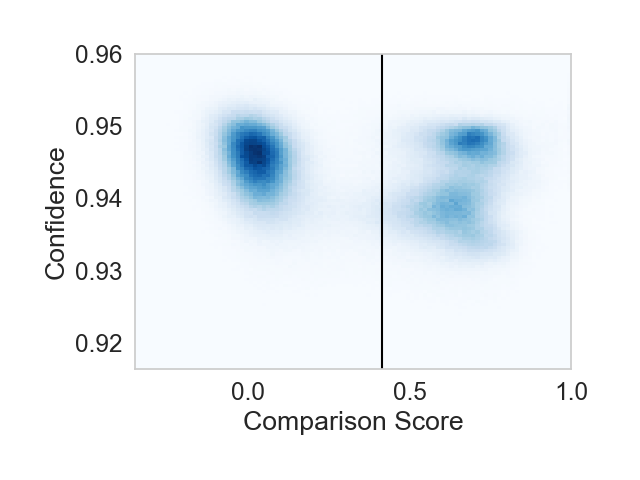}}
     \end{subfigure}
     \begin{subfigure}[MF, $\alpha = 2$\label{intconf}]{ %
     \includegraphics[clip, width=0.18\textwidth]{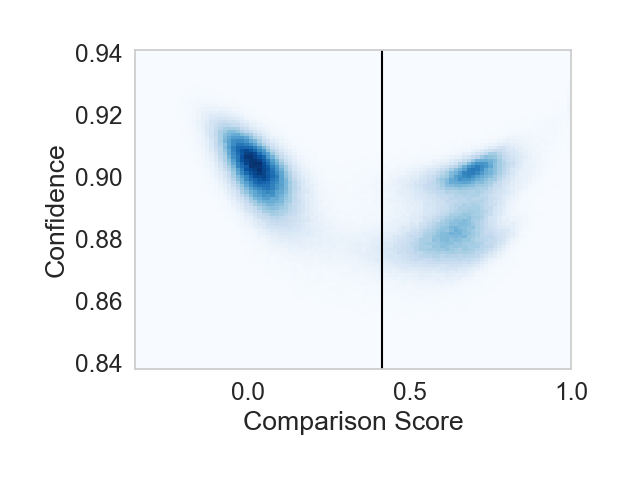}}
     \end{subfigure}
     \begin{subfigure}[MF, $\alpha = 3$\label{intconf}]{ %
     \includegraphics[clip, width=0.18\textwidth]{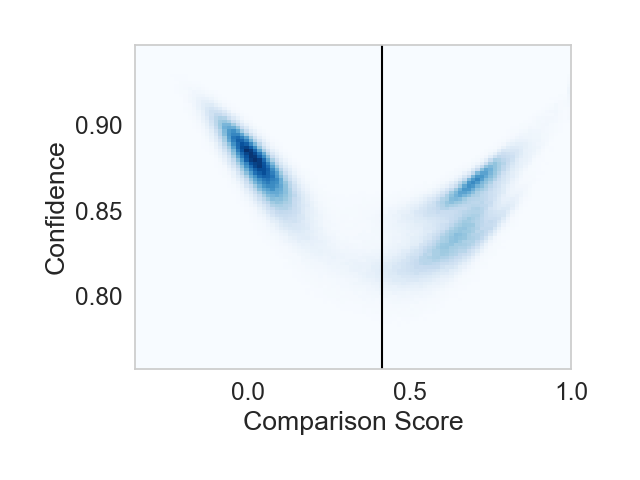}}
     \end{subfigure}
     \begin{subfigure}[MF, $\alpha = 5$\label{intconf}]{ %
     \includegraphics[clip, width=0.18\textwidth]{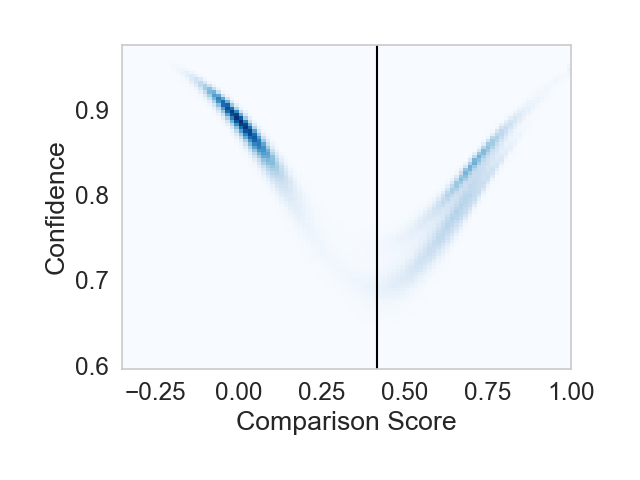}}
     \end{subfigure}
     \begin{subfigure}[MF, $\alpha = 7$\label{intconf}]{ %
     \includegraphics[clip, width=0.18\textwidth]{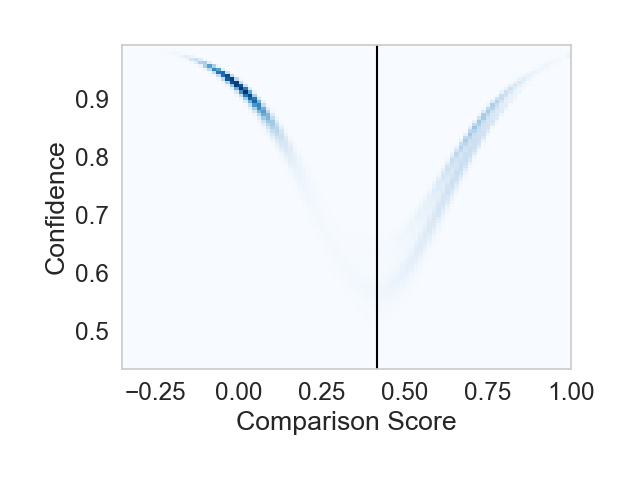}}
     \end{subfigure}
     
     \begin{subfigure}[CF, $\alpha = 1$\label{intconf}]{ %
     \includegraphics[clip, width=0.18\textwidth]{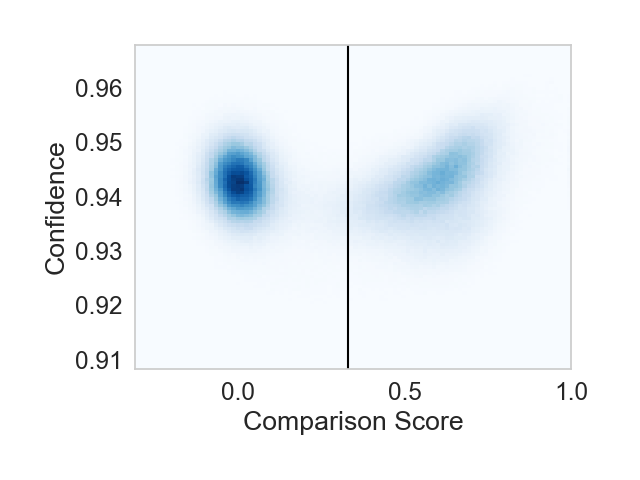}}
     \end{subfigure}
     \begin{subfigure}[CF, $\alpha = 2$\label{intconf}]{ %
     \includegraphics[clip, width=0.18\textwidth]{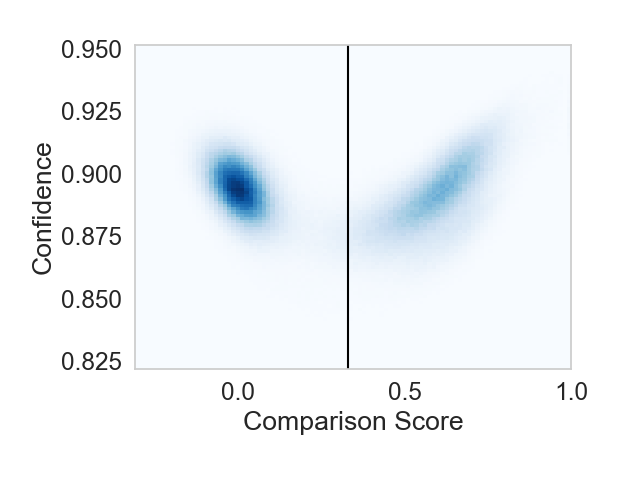}}
     \end{subfigure}
     \begin{subfigure}[CF, $\alpha = 3$\label{intconf}]{ %
     \includegraphics[clip, width=0.18\textwidth]{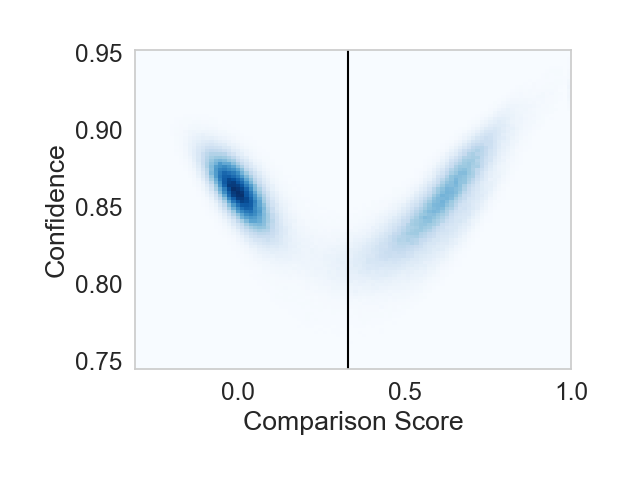}}
     \end{subfigure}
     \begin{subfigure}[CF, $\alpha = 5$\label{intconf}]{ %
     \includegraphics[clip, width=0.18\textwidth]{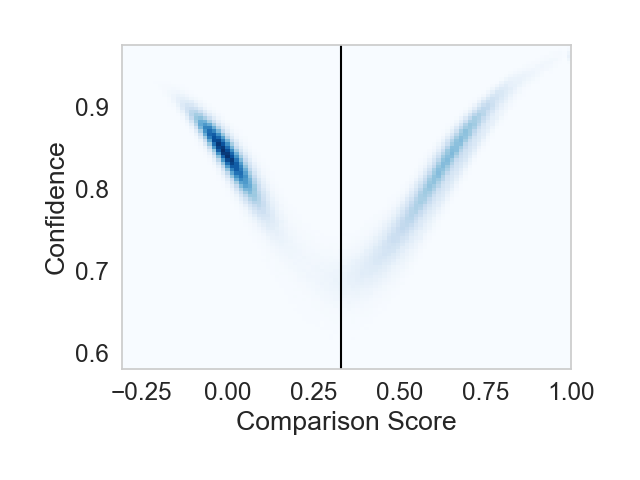}}
     \end{subfigure}
     \begin{subfigure}[CF, $\alpha = 7$\label{intconf}]{ %
     \includegraphics[clip, width=0.18\textwidth]{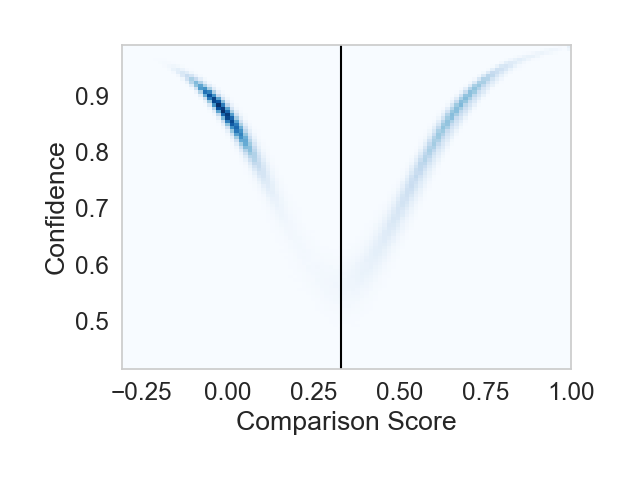}}
     \end{subfigure}

    \caption{\textbf{Influence of the $\alpha$-parameter on the Decision Confidence on the Adience dataset:} Heatmaps based on the decision confidence and comparison score. With a higher $\alpha$-value, the variety in confidence increases, and the distribution of the values gets more shaped leading to a lower confidence around the decision threshold (black asymptote).}
    \label{heat_adience}
\end{figure*}

In this section, we briefly revisit the $\alpha$-parameter used in the decision confidence function (Equation 5 and 6 in the paper) for the trade-off between intuitive decision confidence and uncertainty. Figure \ref{heat_adience} shows the heatmaps for different $\alpha$ values on the Adience dataset. With an increase in $\alpha$, the range of decision confidence values increases, and the distribution gets more shaped in a way that assigns lower confidence values near the decision threshold (black asymptote). The two different confidence clusters that can be determined on the genuine side of the decision threshold are probably due to biased groups, such as children, that are present in the Adience dataset. 

\begin{figure*}
   \centering
    \begin{subfigure}[AF, $\alpha = 1$\label{intconf}]{ %
     \includegraphics[clip, width=0.18\textwidth]{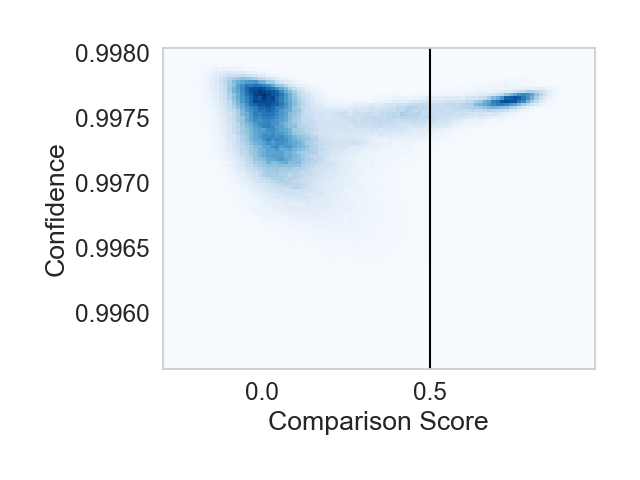}}
     \end{subfigure}
     \begin{subfigure}[AF, $\alpha = 2$\label{intconf}]{ %
     \includegraphics[clip, width=0.18\textwidth]{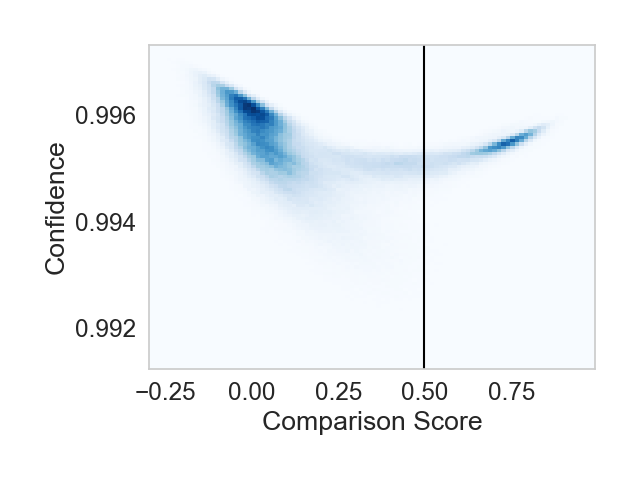}}
     \end{subfigure}
     \begin{subfigure}[AF, $\alpha = 3$\label{intconf}]{ %
     \includegraphics[clip, width=0.18\textwidth]{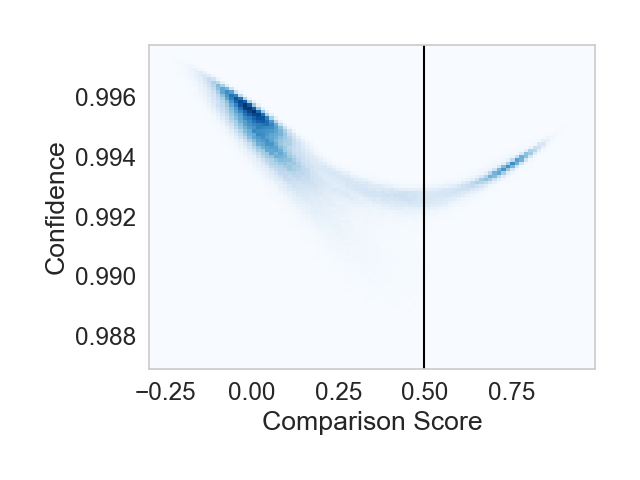}}
     \end{subfigure}
     \begin{subfigure}[AF, $\alpha = 5$\label{intconf}]{ %
     \includegraphics[clip, width=0.18\textwidth]{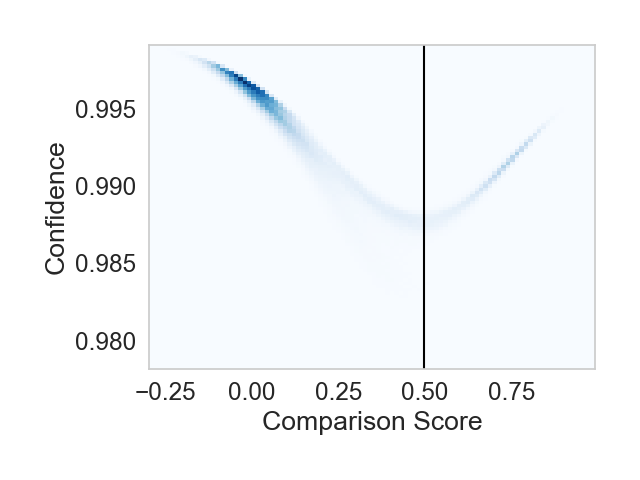}}
     \end{subfigure}
     \begin{subfigure}[AF, $\alpha = 7$\label{intconf}]{ %
     \includegraphics[clip, width=0.18\textwidth]{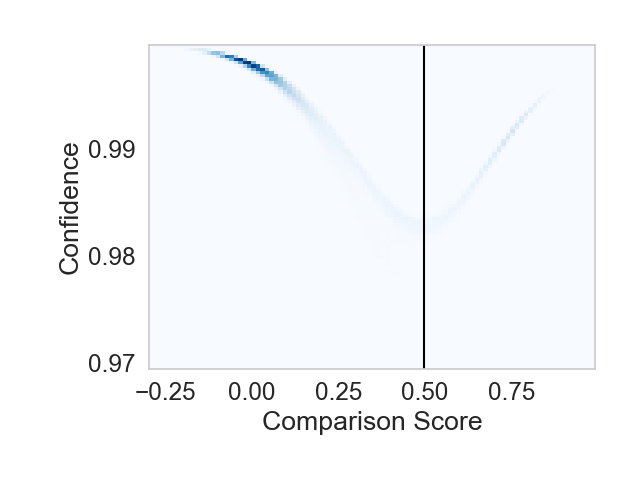}}
     \end{subfigure}
     
     \begin{subfigure}[MF, $\alpha = 1$\label{intconf}]{ %
     \includegraphics[clip, width=0.18\textwidth]{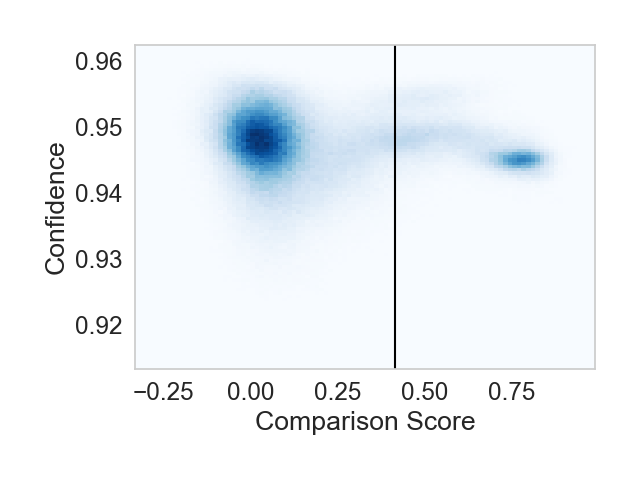}}
     \end{subfigure}
     \begin{subfigure}[MF, $\alpha = 2$\label{intconf}]{ %
     \includegraphics[clip, width=0.18\textwidth]{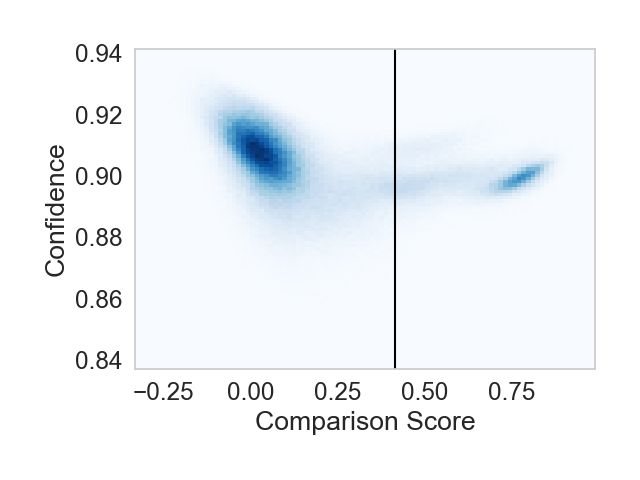}}
     \end{subfigure}
     \begin{subfigure}[MF, $\alpha = 3$\label{intconf}]{ %
     \includegraphics[clip, width=0.18\textwidth]{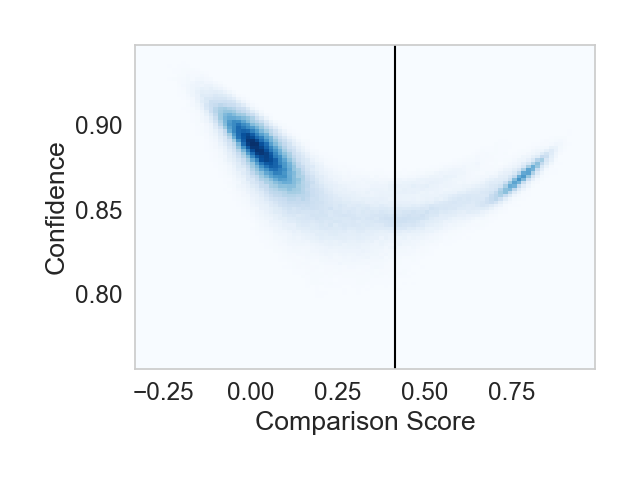}}
     \end{subfigure}
     \begin{subfigure}[MF, $\alpha = 5$\label{intconf}]{ %
     \includegraphics[clip, width=0.18\textwidth]{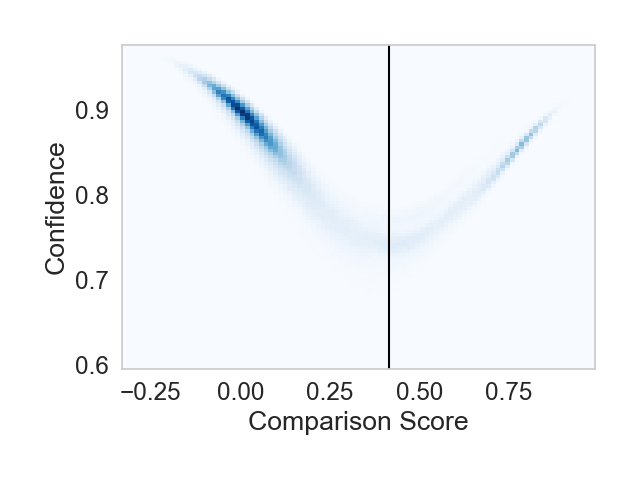}}
     \end{subfigure}
     \begin{subfigure}[MF, $\alpha = 7$\label{intconf}]{ %
     \includegraphics[clip, width=0.18\textwidth]{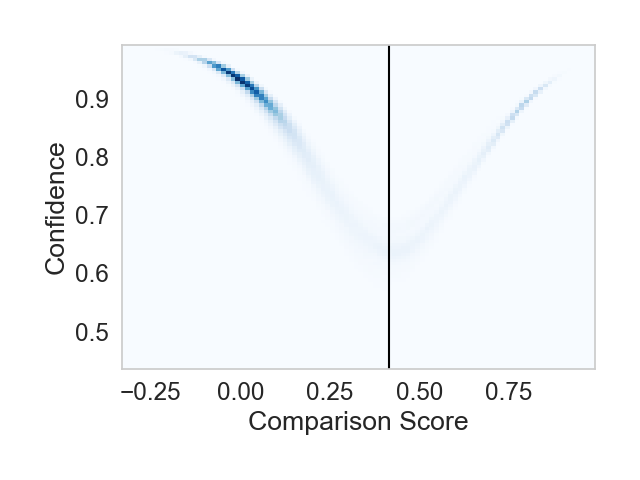}}
     \end{subfigure}
     
     \begin{subfigure}[CF, $\alpha = 1$\label{intconf}]{ %
     \includegraphics[clip, width=0.18\textwidth]{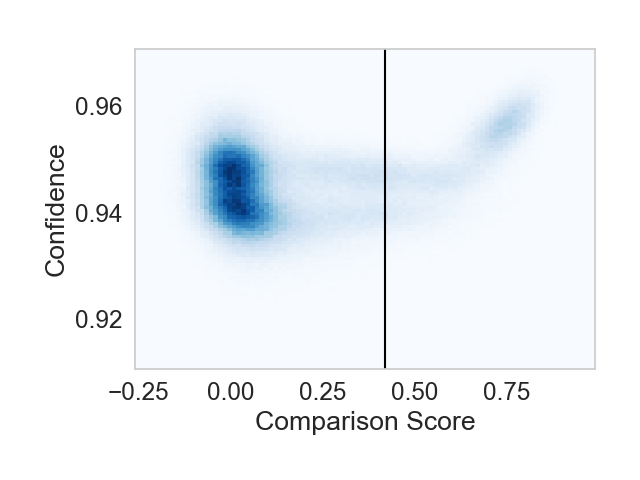}}
     \end{subfigure}
     \begin{subfigure}[CF, $\alpha = 2$\label{intconf}]{ %
     \includegraphics[clip, width=0.18\textwidth]{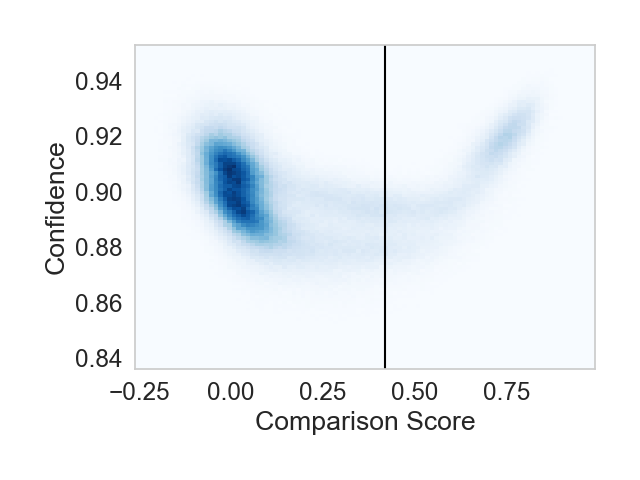}}
     \end{subfigure}
     \begin{subfigure}[CF, $\alpha = 3$\label{intconf}]{ %
     \includegraphics[clip, width=0.18\textwidth]{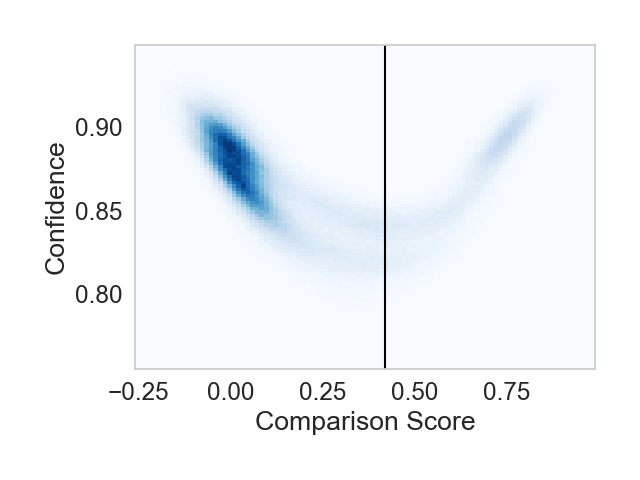}}
     \end{subfigure}
     \begin{subfigure}[CF, $\alpha = 5$\label{intconf}]{ %
     \includegraphics[clip, width=0.18\textwidth]{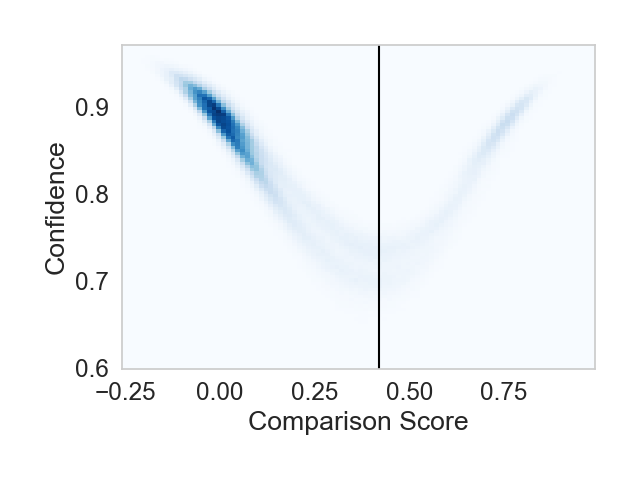}}
     \end{subfigure}
     \begin{subfigure}[CF, $\alpha = 7$\label{intconf}]{ %
     \includegraphics[clip, width=0.18\textwidth]{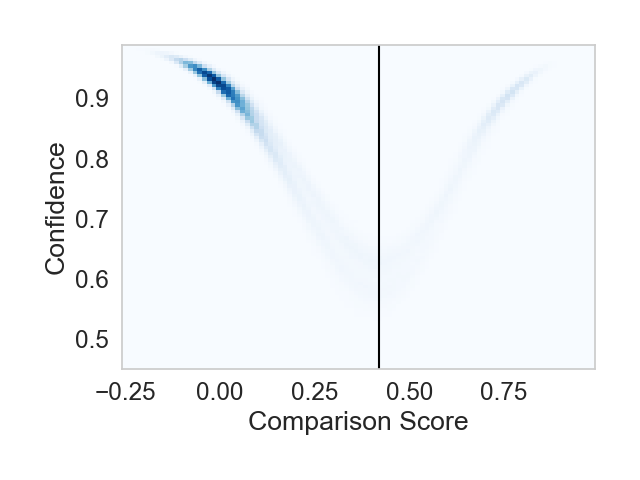}}
     \end{subfigure}
    \caption{\textbf{Influence of the $\alpha$-parameter on the Decision Confidence on the XQLFW dataset:} Heatmaps based on the confidence and comparison score. Similarly to the experiments on the Adience dataset, it can be observed that with a higher $\alpha$-value, the variety in confidence increases. With an increase in $\alpha$, the lowest decision confidence is more and more shifted to the decision threshold.}
   \label{heat_xqlfw}
\end{figure*}

Similar heatmaps are shown in Figure \ref{heat_xqlfw} for the experiments performed on the XQLFW dataset. Due to the cross-quality pairs, the distributions are less separated than in the Adience dataset, which can be also observed in the heatmaps. Furthermore, we also investigated low-quality pairs that lead to high variation in the confidence for small comparison scores.

\begin{figure*}
    \centering
    \begin{subfigure}[ArcFace\label{intconf}]{ %
     \includegraphics[clip, width=0.5\textwidth]{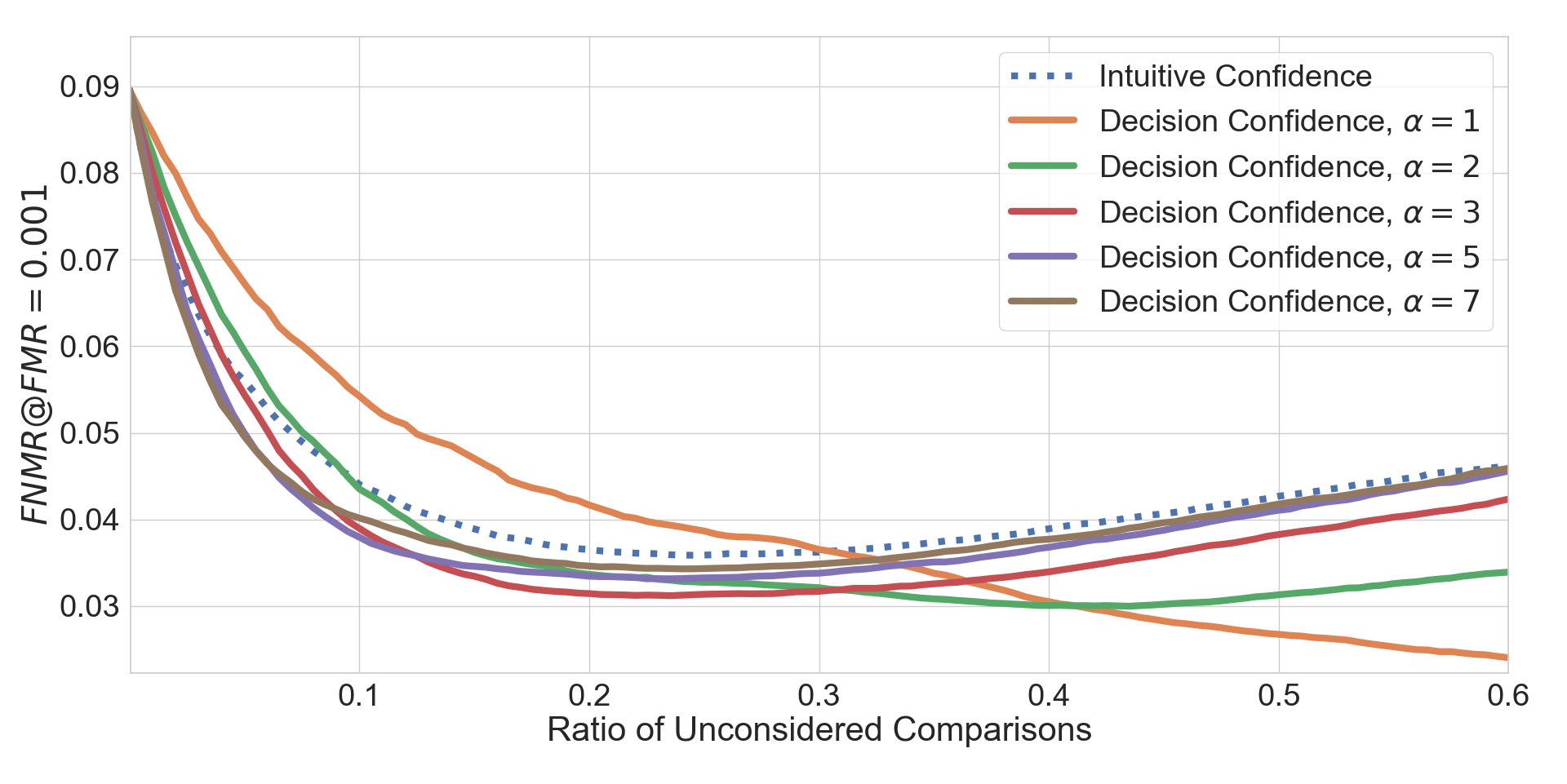}}
    \end{subfigure}
     \begin{subfigure}[MagFace\label{intconf}]{ %
     \includegraphics[clip, width=0.5\textwidth]{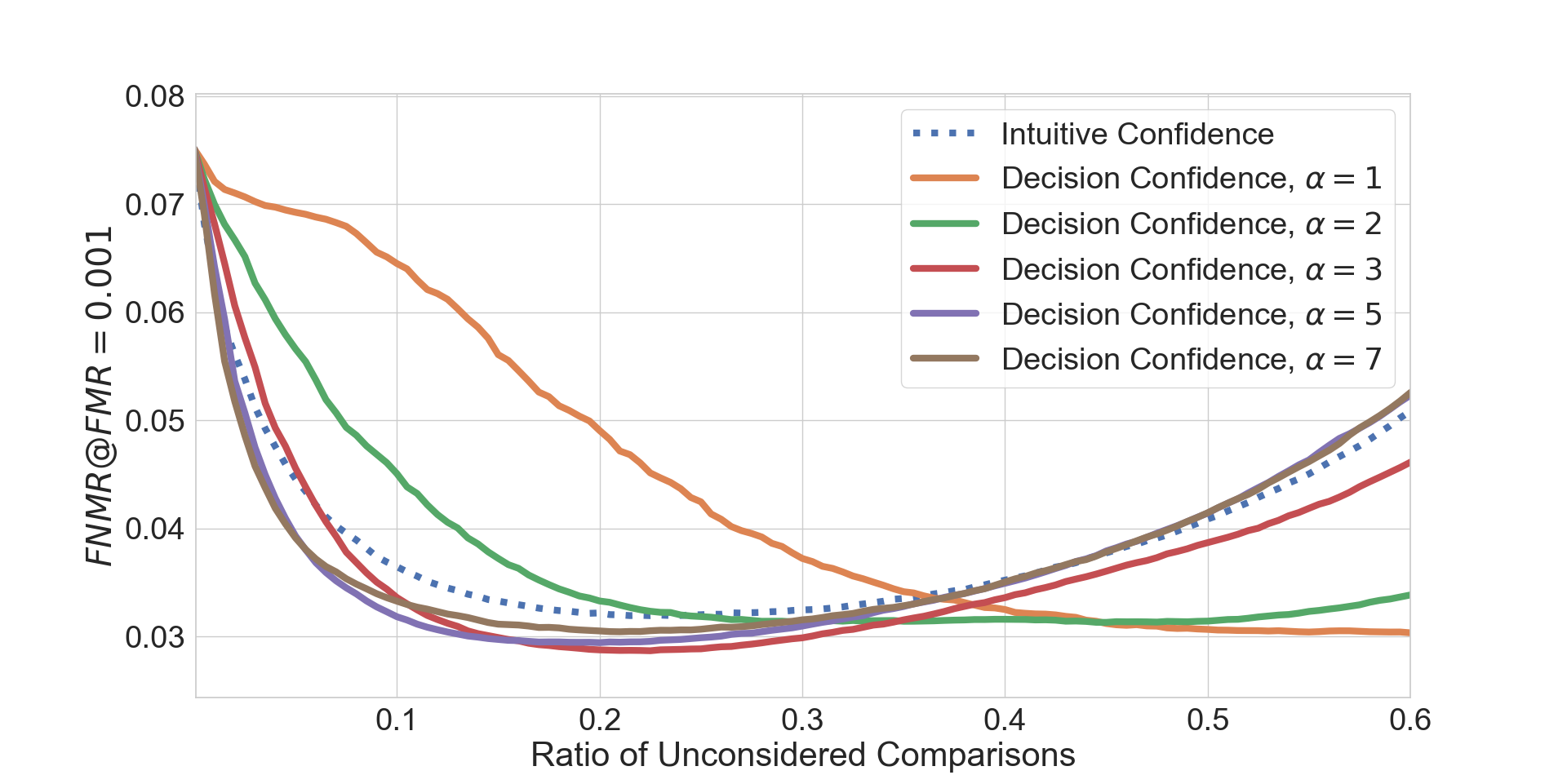}}
     \end{subfigure}
    \begin{subfigure}[CurricularFace\label{intconf}]{ %
     \includegraphics[clip, width=0.5\textwidth]{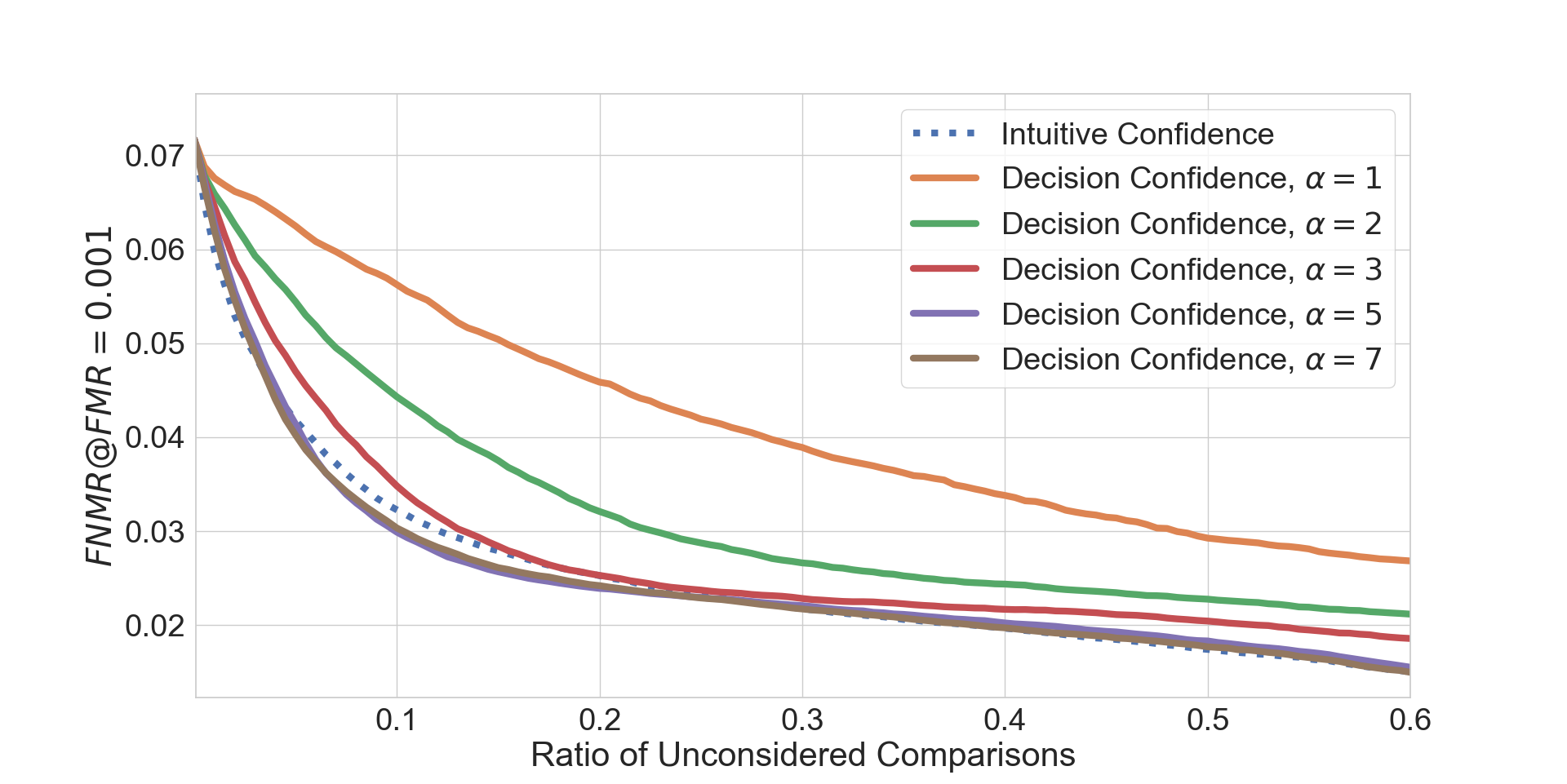}}
    \end{subfigure}
    \caption{\textbf{Influence of the $\alpha$-parameter on the Error-vs-Reject curve on the Adience dataset:} Comparisons are removed depending on their calculated confidence and the performance is evaluated on the remaining comparisons. The $\alpha$-parameter determines the trade-off between intuitive confidence and uncertainty. $\alpha=3$ and $\alpha=5$ outperform the intuitive confidence baseline on all models.}
    \label{adience}
\end{figure*}

\begin{figure*}
    \centering
    \begin{subfigure}[ArcFace\label{intconf}]{ %
     \includegraphics[clip, width=0.5\textwidth]{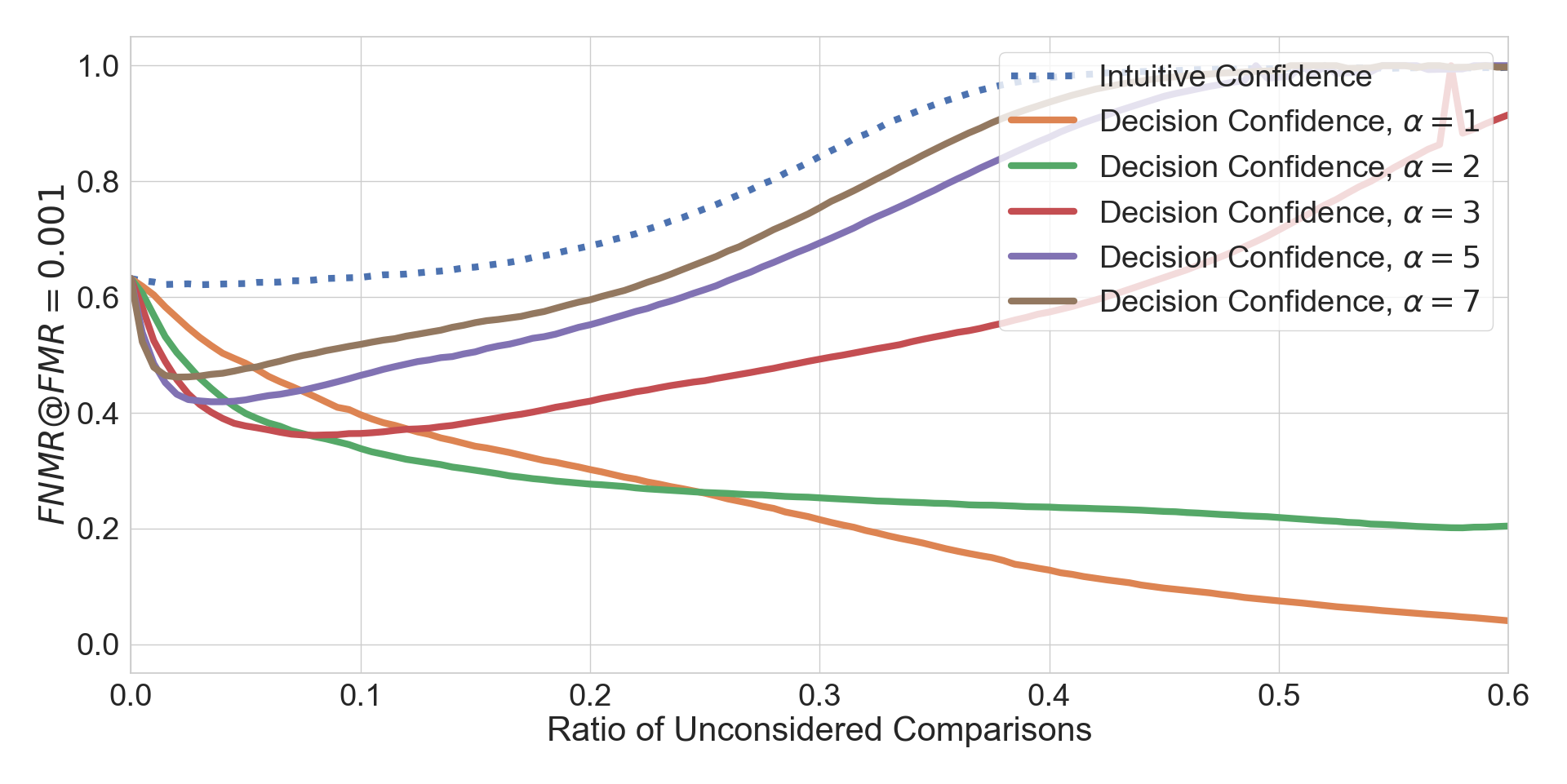}}
    \end{subfigure}
    \begin{subfigure}[MagFace\label{intconf}]{ %
     \includegraphics[clip, width=0.5\textwidth]{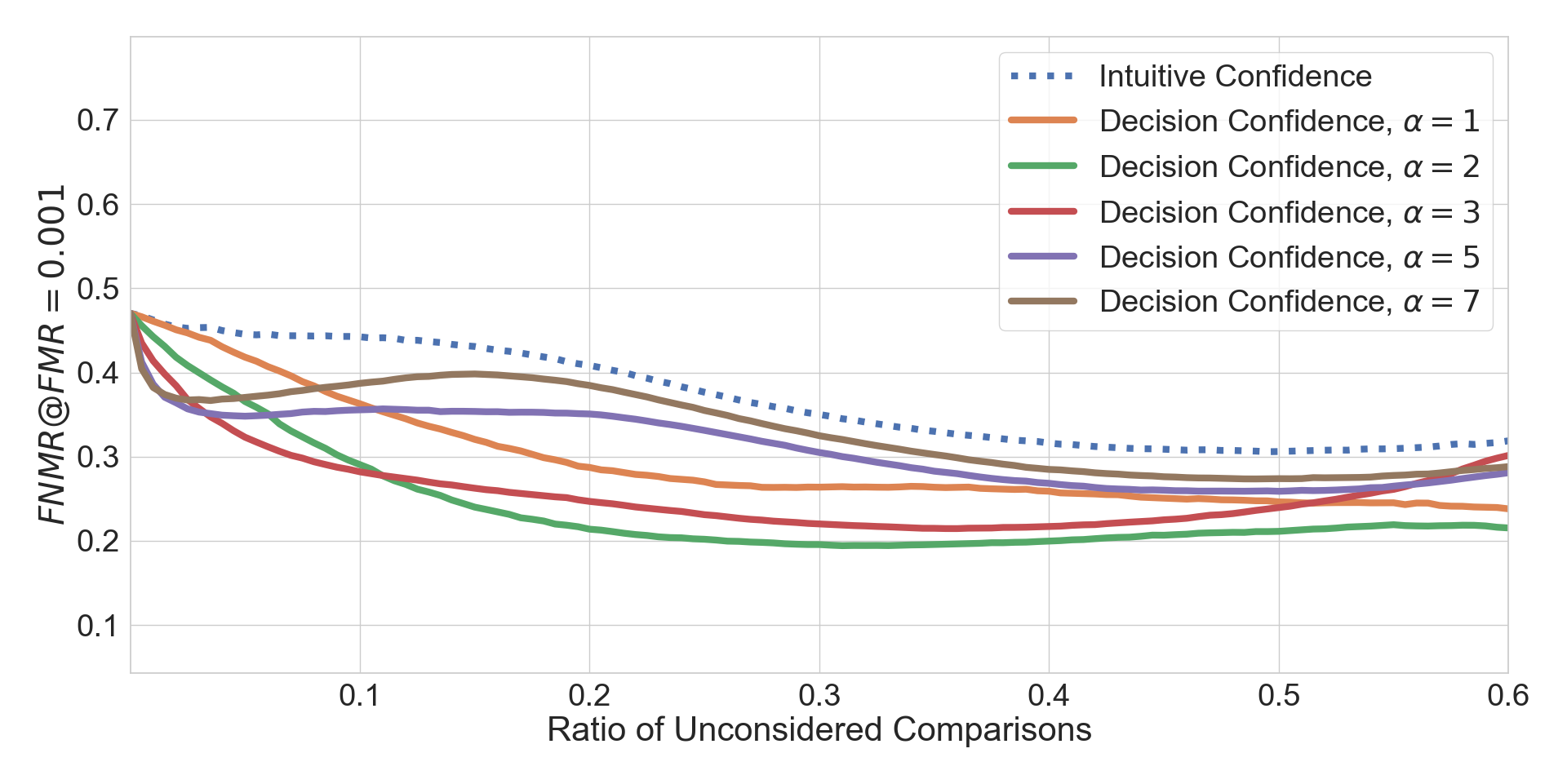}}
    \end{subfigure}
    \begin{subfigure}[CurricularFace\label{intconf}]{ %
     \includegraphics[clip, width=0.5\textwidth]{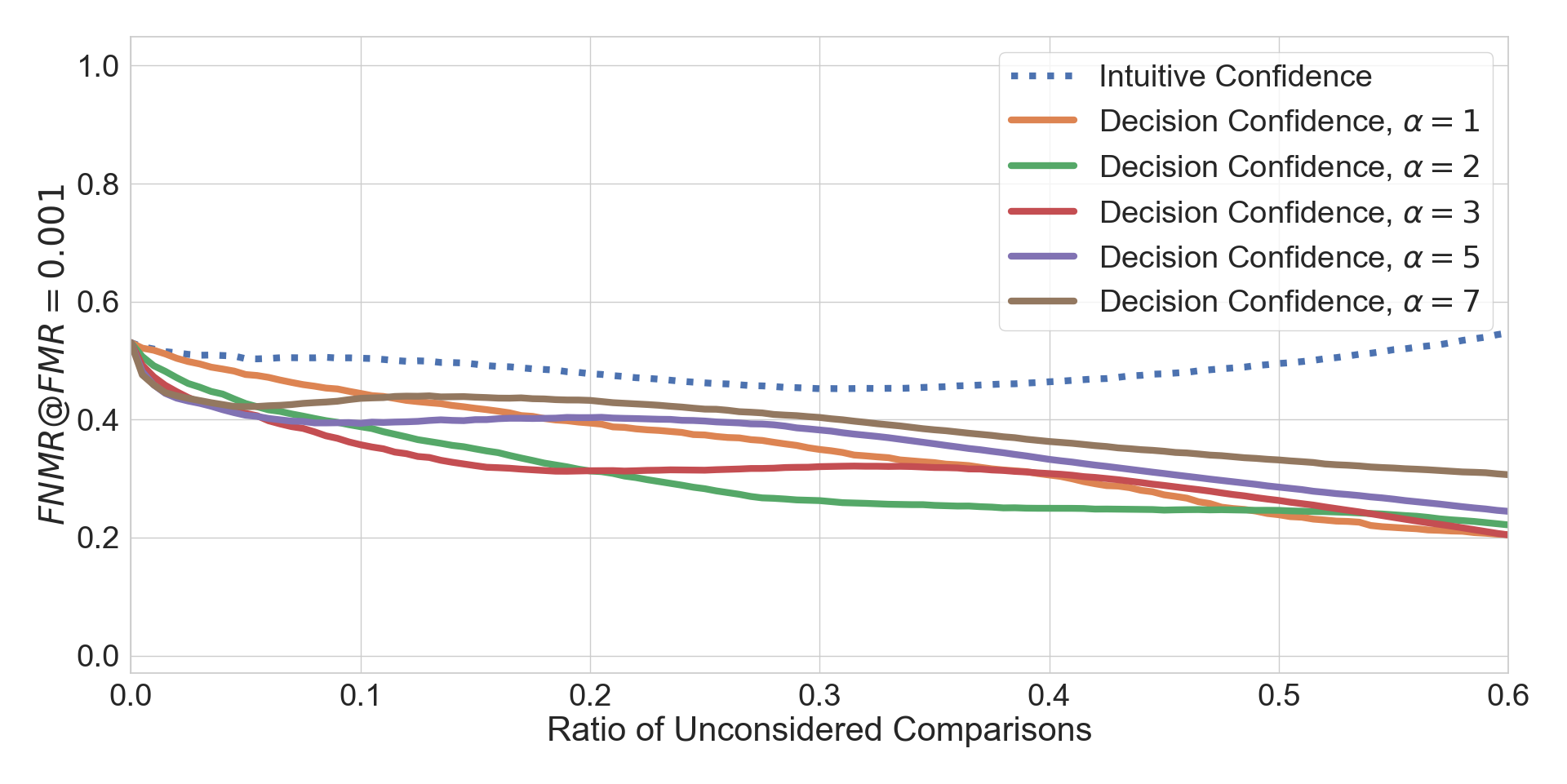}}
    \end{subfigure}
    \caption{\textbf{Influence of the $\alpha$-parameter on the Error-vs-Reject curve on the XQLFW dataset:} Comparisons are removed depending on their calculated confidence and the performance is evaluated on the remaining comparisons. The proposed decision confidence outperforms the intuitive confidence baseline on the difficult XQLFW dataset. The decrease in performance (the increase of the error) is due to the imbalance of the data at the operational point of FNMR@FMR=0.001.}
    \label{xqlfw}
\end{figure*}

The influence of the $\alpha$-parameter on the error-vs-reject curves is shown in Figure \ref{adience} on the Adience dataset and in Figure \ref{xqlfw} on the XQLFW dataset. Since at the selected threshold of FNMR@FMR=0.001 the genuine and imposter comparisons are unevenly distributed around the threshold, the recognition error increased when the ratio of unconsidered comparisons become higher. However, in most cases, the proposed decision confidence outperforms the threshold-based intuitive confidence baseline. 

\newpage
\bibliography{egbib}